\pdfoutput=1
\documentclass[11pt]{article}

\usepackage[preprint]{acl}

\usepackage{times}
\usepackage{latexsym}

\usepackage[T1]{fontenc}
\usepackage[utf8]{inputenc}

\usepackage{microtype}

\usepackage{inconsolata}

\usepackage{graphicx}
\usepackage{bbold}
\usepackage{color}
\usepackage{colortbl}

\title{Enhancing Healthcare LLM Trust with Atypical Presentations Recalibration}

\author{
  \textbf{Jeremy Qin\textsuperscript{1,2}},
  \textbf{Bang Liu\textsuperscript{1,2}},
  \textbf{Quoc Dinh Nguyen\textsuperscript{1,3}}\\
  \\
  \textsuperscript{1}Université de Montréal,
  \textsuperscript{2}Mila,
  \textsuperscript{3}CRCHUM\\
  \small{
    \textbf{Correspondence:} \href{mailto:jeremy.qin@umontreal.ca}{jeremy.qin@umontreal.ca}, \href{mailto:bang.liu@umontreal.ca}{bang.liu@umontreal.ca}, \href{mailto:quoc.dinh.nguyen@umontreal.ca}{quoc.dinh.nguyen@umontreal.ca}
  }
}

\begin{document}
\maketitle
% \begin{abstract}
% This document is a supplement to the general instructions for *ACL authors. It contains instructions for using the \LaTeX{} style files for ACL conferences.
% The document itself conforms to its own specifications, and is therefore an example of what your manuscript should look like.
% These instructions should be used both for papers submitted for review and for final versions of accepted papers.
% \end{abstract}

\begin{abstract}
% Black-box large language models (LLMs) are rapidly being deployed in various environments. It then becomes crucial for these models to effectively convey their confidence and uncertainty, particularly in high-stakes settings. However, these models often exhibit overconfidence, leading to potential risks and misjudgments. Efforts on exploring techniques to elicit and calibrate the confidence levels of LLMs have primarily focused on general reasoning datasets and have achieved only modest improvements in calibration. Achieving accurate calibration is crucial in order to facilitate decision-making and prevent adverse outcomes. However, it is particularly challenging due to the inherent complexity and variability of the tasks these models perform. In this work, we investigate the miscalibration behavior of black-box LLMs specifically within the healthcare setting. We propose a novel method, \textit{Atypical Presentations Aware Recalibration}, which leverages the concept of atypical presentations to adjust the model's confidence estimates. We demonstrate that incorporating the concept of atypicality brings significant improvements in calibration, with an observed reduction in calibration errors of around 60\% and outperforming all others methods. Furthermore, we provide an in-depth analysis of how atypicality sits inside the recalibration framework.

Black-box large language models (LLMs) are increasingly deployed in various environments, making it essential for these models to effectively convey their confidence and uncertainty, especially in high-stakes settings. However, these models often exhibit overconfidence, leading to potential risks and misjudgments. Existing techniques for eliciting and calibrating LLM confidence have primarily focused on general reasoning datasets, yielding only modest improvements. Accurate calibration is crucial for informed decision-making and preventing adverse outcomes but remains challenging due to the complexity and variability of tasks these models perform. In this work, we investigate the miscalibration behavior of black-box LLMs within the healthcare setting. We propose a novel method, \textit{Atypical Presentations Recalibration}, which leverages atypical presentations to adjust the model's confidence estimates. Our approach significantly improves calibration, reducing calibration errors by approximately 60\% on three medical question answering datasets and outperforming existing methods such as vanilla verbalized confidence, CoT verbalized confidence and others. Additionally, we provide an in-depth analysis of the role of atypicality within the recalibration framework. The code can be found at \url{https://github.com/jeremy-qin/medical_confidence_elicitation}
\end{abstract}

% \section{Introduction}
\section{Introduction}
\begin{figure}[ht]
  \centering
  \includegraphics[width=1\columnwidth]{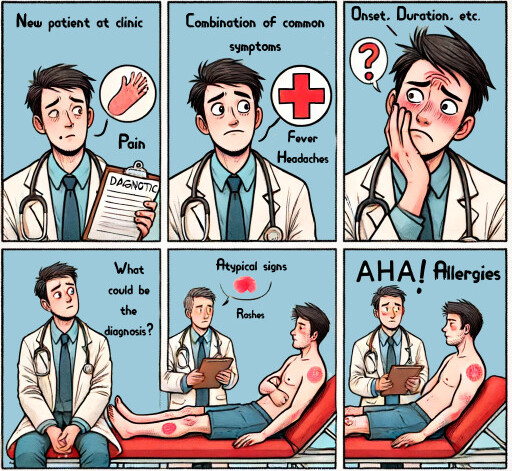} \hfill
  \caption{A physician diagnoses a patient who returned from a camping trip, presenting a combination of common symptoms and signs like fever and headaches. However, by recognizing rashes an atypical symptom, the physician ultimately identifies the condition as an allergy.}
  \label{fig:comic}
\end{figure}

Despite recent successes and innovations in large language models (LLMs), their translational value in high-stakes environments, such as healthcare, has not been fully realized. This is primarily due to concerns about the trustworthiness and transparency of these models, stemming from their complex architecture and black-box nature. Recent studies \citep{xiong2024llms, shrivastava2023llamas, tian2023just, he2023investigating, rivera2024combining, chen2023quantifying} have begun to explore methods for eliciting confidence and uncertainty estimates from these models in order to enhance trustworthiness and transparency. The ability to convey uncertainty and confidence is central to clinical medicine \citep{Banerji2023} and plays a crucial role in facilitating rational and informed decision-making. This underscores the importance of investigating and utilizing calibrated confidence estimates for the medical domain.

Previous work on confidence elicitation and calibration of large language models (LLMs) has mainly focused on general reasoning and general knowledge datasets for tasks such as logical reasoning, commonsense reasoning, mathematical reasoning, and scientific knowledge \citep{kuhn2023semantic, xiong2024llms, tian2023just, tanneru2023quantifying, chen2023quantifying}. Few studies have investigated tasks that require expert knowledge, and these have shown considerable room for improvement. Moreover, with the success of many closed-source LLMs, such as GPT-3.5 and GPT-4, which do not allow access to token-likelihoods and text embeddings, it has become prevalent to develop tailored methods for eliciting confidence estimates. However, most approaches developed consist of general prompting and sampling strategies without using domain-specific characteristics.

Traditionally, clinicians are taught to recognize and diagnose typical presentations of common illnesses based on patient demographics, symptoms and signs, test results, and other standard indicators \citep{Harada2024}. However, the frequent occurrence of atypical presentations is often overlooked \citep{1180014634}. Failing to identify atypical presentations can result in worse outcomes, missed diagnoses, and lost opportunities for treating common conditions. Thus, awareness of atypical presentations in clinical practice is fundamental to providing high-quality care and making informed decisions. Figure \ref{fig:comic} depicts a simplistic example of how atypicality plays a role in diagnosis. Incorporating the concept of atypicality has been shown to improve uncertainty quantification and model performance for discriminative neural networks and white-box large language models \citep{yuksekgonul2023confidence}. This underscores the importance of leveraging atypical presentations to enhance the calibration of LLMs, particularly in high-stakes environments like healthcare.

Our study aims to address these gaps by first investigating the miscalibration of black-box LLMs when answering medical questions using non-logit-based uncertainty quantification methods. We begin by testing various baseline methods to benchmark the calibration of these models across a range of medical question-answering datasets. This benchmarking provides a comprehensive understanding of the current state of calibration in LLMs within the healthcare domain and highlights the limitations of existing approaches.

Next, we propose a new recalibration framework based on the concept of atypicality, termed \textbf{Atypical Presentations Recalibration}. This method leverages atypical presentations to adjust the model's confidence estimates, making them more accurate and reliable. Under this framework, we construct two distinct atypicality-aware prompting strategies for the LLMs, encouraging them to consider and reason over atypical cases explicitly. We then compare the performance and calibration of these strategies against the baseline methods to evaluate their effectiveness.

Finally, our empirical results reveal several key findings. First, black-box LLMs often fail to provide calibrated confidence estimates when answering medical questions and tend to remain overconfident. Second, our proposed Atypical Presentations Aware Recalibration method significantly improves calibration, reducing calibration errors by approximately 60\% on three medical question answering datasets and consistently outperforming existing baseline methods across all datasets. Third, we observe that atypicality interacts in a complex manner with both performance and calibration, suggesting that considering atypical presentations is crucial for developing more accurate and trustworthy LLMs in healthcare settings.

% These instructions are for authors submitting papers to *ACL conferences using \LaTeX. They are not self-contained. All authors must follow the general instructions for *ACL proceedings,\footnote{\url{https://github.com/jeremy-qin/medical_confidence_elicitation}} and this document contains additional instructions for the \LaTeX{} style files.

% The templates include the \LaTeX{} source of this document (\texttt{acl\_latex.tex}),
% the \LaTeX{} style file used to format it (\texttt{acl.sty}),
% an ACL bibliography style (\texttt{acl\_natbib.bst}),
% an example bibliography (\texttt{custom.bib}),
% and the bibliography for the ACL Anthology (\texttt{anthology.bib}).

\section{Background and Related Work}
\subsection{Confidence and Uncertainty quantification in LLMs}
Confidence and uncertainty quantification is a well-established field, but the recent emergence of large language models (LLMs) has introduced new challenges and opportunities. Although studies have shown a distinction between confidence and uncertainty, we will use these terms interchangeably in our work.

Research on this topic can be broadly categorized into two areas: approaches targeting closed-source models and those focusing on open-source models. The growing applications of commercial LLMs, due to their ease of use, have necessitated particular methods to quantify their confidence. For black-box LLMs, a natural approach is to prompt them to express confidence verbally, a method known as verbalized confidence, first introduced by \citet{lin2022teaching}. Other studies have explored this approach specifically for language models fine-tuned with reinforcement learning from human feedback (RLHF) \citep{tian2023just, he2023investigating}. Additionally, some research has proposed new metrics to quantify uncertainty \citep{ye2024benchmarking, tanneru2023quantifying}.

Our work aligns most closely with \citet{xiong2024llms}, who presented a framework that combines prompting strategies, sampling techniques, and aggregation methods to elicit calibrated confidences from LLMs. While previous studies primarily benchmarked their methods on general reasoning tasks, our study focuses on the medical domain, where accurate uncertainty quantification is critical for diagnosis and decision-making. We evaluate LLM calibration using the framework defined by \citet{xiong2024llms} and propose a framework consisting of a new prompting strategy and aggregation method, termed \textit{Atypicality Presentations Recalibration}, which shows significant improvements in calibrating LLM uncertainty in the medical domain.

\begin{table*}[]
\begin{tabular}{p{0.3\textwidth} p{0.7\textwidth}}
\hline
\textbf{Method}        & \textbf{Prompt Template}                                                                                                                                                                                                                                                \\ \hline
Vanilla                & Read the following question. Provide your answer and your confidence level (0\% to 100\%).                                                                                                                                                                       \\ \hline
Atypical Scenario      & Read the following question. Assess the atypicality of the scenario described with a score between 0 and 1 with 0 being highly atypical and 1 being typical. Provide your answer, atypicality score and confidence level.                                        \\ \hline
Atypical Presentations & Read the following question. Assess each symptom and signs with respect to its typicality in the described scenario with a score between 0 and 1 with 0 being highly atypical and 1 being typical. Provide your answer, atypicality scores and confidence level. \\ \hline
\end{tabular}
\caption{\label{tab:prompt_templates}
    Illustrations of the vanilla prompting and Atypical Presentations Aware Recalibration prompting strategies (complete prompts in Appendix \ref{sec:appendixB})
  }
\end{table*}

\subsection{Atypical Presentations}
Atypical presentations have garnered increasing attention and recognition in the medical field due to their critical role in reducing diagnostic errors and enhancing problem-based learning in medical education \citep{VONNES2021458, Kostopoulou2008, Matulis2020, Bai2023}. Atypical presentations are defined as "a shortage of prototypical features most frequently encountered in patients with the disease, features encountered in advanced stages of the disease, or features commonly listed in medical textbooks" \citep{Kostopoulou2008, Harada2024}. This concept is particularly important in geriatrics, where older patients often present atypically, and in medical education, where it prompts students to engage in deeper reflection during diagnosis.

Given the increasing emphasis on atypical presentations in medical decision-making, it is pertinent to explore whether this concept can be leveraged to calibrate machine learning models. \citet{yuksekgonul2023confidence} were the first to incorporate atypicality into model calibration for classification tasks. Our work extends this approach to generative models like LLMs, integrating atypical presentations to achieve more accurate and calibrated confidence estimates.

% \begin{figure*}[t]
%   \includegraphics[width=1\linewidth]{latex/images/APAF.drawio.png} \hfill
%   \caption {Overview of our Atypical Presentation Aware Calibration Framework}
%   \label{fig:framework}
% \end{figure*}

\section{Method}
In this section, we describe the methods used to elicit confidence from large language models (LLMs) as well as our recalibration methods. Calibration in our settings refers to the alignment between the confidence estimates and the true likelihood of outcomes \citep{yuksekgonul2023confidence, doi:10.1198/016214506000001437}. Our experiments are based on the framework described by \citet{xiong2024llms}, which divides the approaches into three main components: prompting, sampling, and aggregation, and uses it as baselines. In their framework, they leverage common prompting strategies such as vanilla prompting and Chain-of-Thoughts while also leveraging the stochasticity of LLMs. In contrast, we propose an approach, \textit{Atypical Presentation Recalibration}, that retrieves atypicality scores and use them as a recalibration method in order to have more accurate confidence estimates. Our framework is mainly divided into two parts: \textit{Atypicality Prompting} and \textit{Atypicality Recalibration}. We explain how each of the three components are applied to our tasks and how we integrate atypicality to develop hybrid methods that combine these elements.

\subsection{Prompting methods}
Eliciting confidence from LLMs can be achieved through various methods, including natural language expressions, visual representations, and numerical scores \citep{kim2024im}. We refer to these methods collectively as verbalized confidence. While there are trade-offs between these methods, we focus on retrieving numerical confidence estimates for better precision and ease of calibration. We design a set of prompts to elicit confidence estimates from LLMs.

\paragraph{Vanilla Prompting}
The most straightforward way to elicit confidence scores from LLMs is to ask the model to provide a confidence score on a certain scale. We term this method as vanilla prompting. This score is then used to assess calibration.

\paragraph{Chain-of-Thought (CoT)}
Eliciting intermediate and multi-step reasoning through simple prompting has shown improvements in various LLM tasks. By allowing for more reflection and reasoning, this method helps the model express a more informed confidence estimate. We use zero-shot Chain-of-Thought (CoT) \citep{kojima2023large} in our study.

\paragraph{Atypicality Prompting}
% Inspired by the notion of atypical presentations seen in medecine, we wish to enhance the reliability and transparency to the decision-making process and reasoning of LLMs by helping it identify the most relevant symptoms and which ones are outliers. Building on this goal, we construct a prompt that incorporates a systematic approach to assessing the typicality of symptoms and features presented in the question. We refer to this prompting strategy as \textbf{Atypical Presentations Prompt}. However, we also take into account that some medical questions often asked do not necessarily require diagnosis, and we propose another prompting strategy. Inspired by the notion of class atypicality mentionned in \citep{yuksekgonul2023confidence}, we propose a comprehensive approach to decision-making and question answering in uncertain contexts by constructing a prompt that assesses the typicality of the question itself. We name this prompting strategy as \textbf{Atypical Scenario Prompt}. The rationale behind this is that it is natural for people to be more uncertain when asked a question they are not used to or familiar with. The final confidence score is computed by adjusting the initial confidence level through a weighting mechanism that incorporates the aggregated atypicality scores. Drawing from insights in economics and finance, where expert judgments are combined with varying weights and exponential utility functions are used to address risk aversion, we propose a non-linear post-hoc calibration method that incorporates these principles. 

Inspired by the concept of atypical presentations in medicine, we aim to enhance the reliability and transparency of LLM decision-making by incorporating atypicality into the confidence estimation process. We develop two distinct prompting strategies to achieve this goal:
\begin{itemize}
\item \textbf{Atypical Presentations Prompt:} This strategy focuses on identifying and highlighting atypical symptoms and features within the medical data. The prompt is designed to guide the LLM to assess the typicality of each symptom presented in the question. By systematically evaluating which symptoms are atypical, the model can better gauge the uncertainty associated with the diagnosis. For example, the prompt might ask the model to rate the typicality of each symptom on a scale from 0 to 1, where 1 represents a typical symptom and 0 represents an atypical symptom. In the following sections of the paper, we will refer to these scores as atypicality scores where the lower the score is the more atypical it is. This information is then used to adjust the confidence score accordingly.
\item \textbf{Atypical Scenario Prompt:} This strategy evaluates the typicality of the question itself. It is based on the notion that questions which are less familiar or more complex may naturally elicit higher uncertainty. The prompt asks the LLM to consider how common or typical the given medical scenario is. For instance, the model might be prompted to rate the overall typicality of the scenario on a similar scale. This approach helps to capture the inherent uncertainty in less familiar or more complex questions.
\end{itemize}

\subsection{Sampling and Aggregation}
\begin{figure*}[ht]
  \includegraphics[width=1\linewidth]{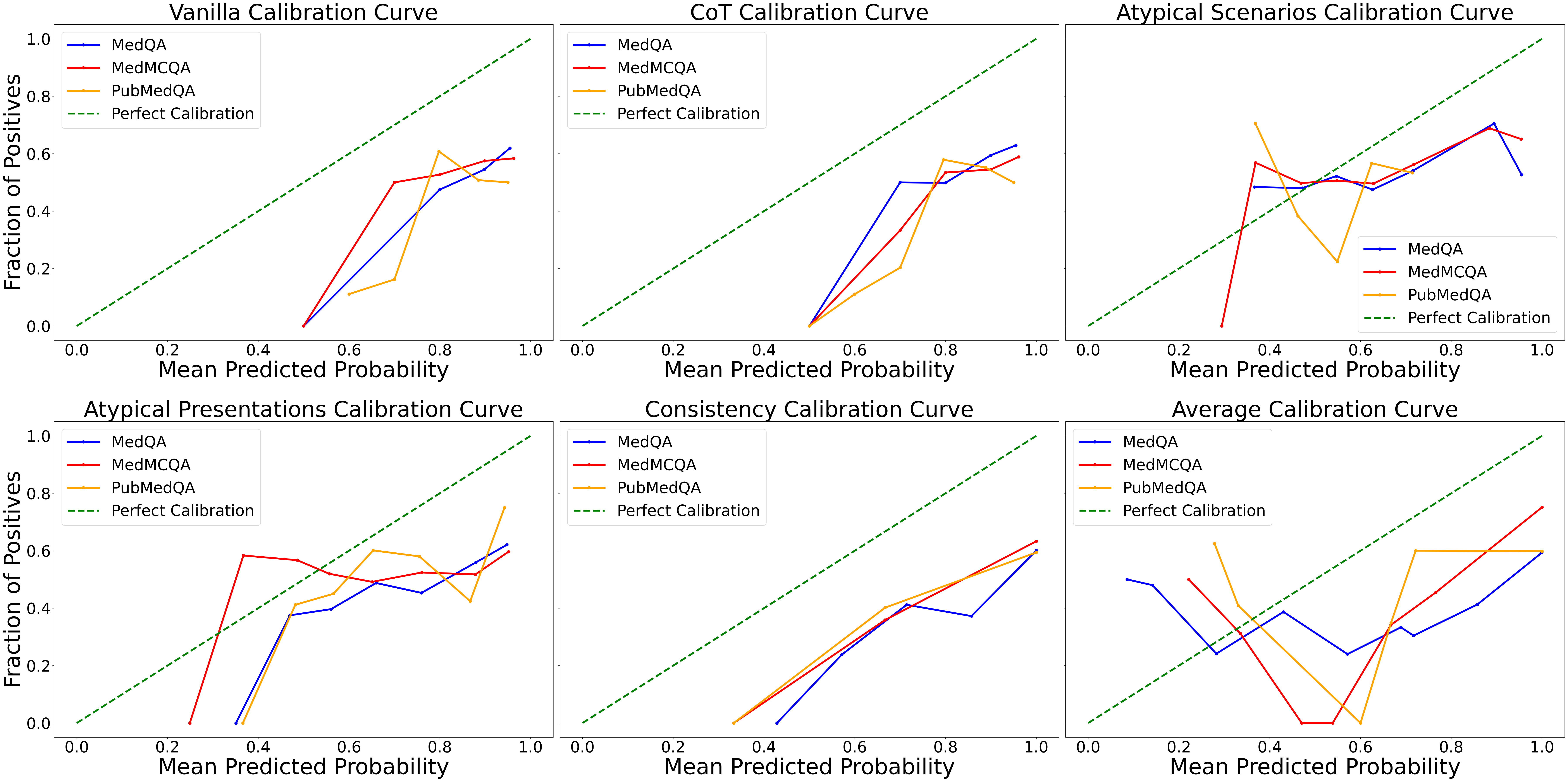} \hfill
  \caption{Calibration Curves of the different methods for GPT-3.5-turbo}
  \label{fig:calibration_curves}
\end{figure*}

While verbalized confidences provide a straightforward way to assess the uncertainty of LLMs, we can also leverage the stochasticity of LLMs \citep{xiong2024llms, rivera2024combining} by generating multiple answers for the same question. Different aggregation strategies can then be used to evaluate how aligned these sampled answers are. We follow the framework defined by \citet{xiong2024llms} for the sampling and aggregation methods and uses them as baselines to our \textit{Atypical Presentations Recalibration} framework.

\paragraph{Self-Random Sampling}
The simplest strategy to generate multiple answers from an LLM is by repeatedly asking the same question and collecting the responses. These responses are then aggregated to produce a final confidence estimate.

\paragraph{Consistency} We use the consistency of agreement between different answers from the LLM as the final confidence estimate \citep{xiong2024llms}. For a given question with a reference answer $\tilde{Y}$, we generate a sample of answers $\hat{Y}_k$. The aggregated confidence $C_{consistency}$ is defined as:
\begin{equation}
C_{consistency} = \frac{1}{K} \sum_{k=1}^K \mathbb{1}\{\hat{Y}_k = \tilde{Y}\}
\end{equation}

\paragraph{Weighted Average}
Building on the consistency aggregation method, we can use a weighting mechanism that incorporates the confidence scores elicited from the LLM. This method weights the agreement between the different answers by their respective confidence scores. The aggregated confidence $C_{average}$ is defined as:
\begin{equation}
    C_{average} = \frac{\sum_{k=1}^K \mathbb{1}\{\hat{Y}_k = \tilde{Y}\} * C_k}{\sum_{k=1}^K C_k}
\end{equation}

\paragraph{Atypicality Recalibration}
To integrate the atypicality scores elicited with \textit{Atypicality Presentations Prompting} into our confidence estimation framework, we propose a non-linear post-hoc recalibration method that combines the initial confidence score with an aggregation of the atypicality assessments. This method draws inspiration from economic and financial models where expert judgments are combined with varying weights and exponential utility functions to address risk aversion. Formally, for an initial confidence $C_i$ of a given question and atypical scores $A_k$, the calibrated confidence $CC_i$ is computed as follows:
\begin{equation}
    CC_i = C_i * \left(\frac{1}{K} \sum_{k=1}^K e^{A_k - 1}\right)
\end{equation}
where $A_k$ takes values in [0,1] and a value of 1 corresponds to a typical value. For the Atypical Scenario Prompt, this equation translates to having K equal to 1. Thus, the final confidence estimate will equal the initial confidence score only if all the atypical scores are 1.

% \paragraph{Hybrid Methods}
% Since the different components are independent, we can combine them to create hybrid methods. By leveraging both prompting strategies and sampling and aggregation strategies together, we can potentially achieve more calibrated confidence estimates. This combination allows us to see if one method can complement the other, resulting in improved calibration.

% \begin{figure}[t]
%   \includegraphics[width=\columnwidth]{images/atypicalprompt.png}
%   \caption{Template for Atypical Presentations Prompt (see exact prompt in Appendix)}
% \end{figure}

\section{Experiments}

\subsection{Experimental Setup}
\paragraph{Datasets} 
Our experiments evaluate the calibration of confidence estimates across three different english medical question-answering datasets. For our experiments, we restricted on evaluating on only the development set of each dataset. \textbf{MedQA} \citep{jin2020disease} consists of 1272 questions based on the United States Medical License Exams and collected from the professional medical board exams. \textbf{MedMCQA} \citep{pmlr-v174-pal22a} is a large-scale multiple-choice question answering dataset with 2816 questions collected from AIIMS \& NEET PG entrance exams covering a wide variety of healthcare topics and medical subjects. \textbf{PubMedQA} \citep{jin2019pubmedqa} is a biomedical question answering dataset with 500 samples collected from PubMed abstracts where the task is to answer research question corresponding to an abstract with yes/no/maybe.

\paragraph{Models} We use a variety of commercial LLMs that includes GPT-3.5-turbo \citep{openai2023gpt35}, GPT-4-turbo \citep{openai2024gpt4}, Claude3-sonnet \citep{anthropic2023claude3} and Gemini 1.0 Pro \citep{google2023gemini}. 

\begin{table*}[t]
\resizebox{1.0\textwidth}{!}{
\begin{tabular}{llllllllllllll}
\hline
               &                                                                                                                                        & \multicolumn{4}{c}{MedQA (n=1272)}                                                                                                                                                                                                                                                                                                                                     & \multicolumn{4}{l}{MedMCQA (n=2816)}                                                                                                                                                                                                                                                                                                                                   & \multicolumn{4}{l}{PubMedQA (n=500)}                                                                                                                                                                                                                                                                                                                                   \\ \hline
Models         & Methods                                                                                                                                & Acc                                                                                   & ECE                                                                                   & Brier                                                                                 & AUC                                                                                   & Acc                                                                                   & ECE                                                                                   & Brier                                                                                 & AUC                                                                                   & Acc                                                                                   & ECE                                                                                    & Brier                                                                                 & AUC                                                                                   \\ \hline
gpt-3.5-turbo  & \begin{tabular}[c]{@{}l@{}}Vanilla\\ CoT\\ Atypical scenario\\ Atypical presentations\\ Consistency (k=3)\\ Average (k=3)\end{tabular} & \begin{tabular}[c]{@{}l@{}}0.526\\ 0.536\\ 0.506\\ 0.506\\ 0.535\\ \textbf{0.539}\end{tabular} & \begin{tabular}[c]{@{}l@{}}0.351\\ 0.318\\ \textbf{0.084}\\ 0.283\\ 0.408\\ 0.398\end{tabular} & \begin{tabular}[c]{@{}l@{}}0.363\\ 0.334\\ \textbf{0.262}\\ 0.332\\ 0.396\\ 0.397\end{tabular} & \begin{tabular}[c]{@{}l@{}}0.553\\ \textbf{0.608}\\ 0.530\\ 0.557\\ 0.567\\ 0.555\end{tabular} & \begin{tabular}[c]{@{}l@{}}0.555\\ 0.525\\ 0.544\\ 0.527\\ \textbf{0.561}\\ \textbf{0.561}\end{tabular} & \begin{tabular}[c]{@{}l@{}}0.323\\ 0.357\\ \textbf{0.128}\\ 0.152\\ 0.350\\ 0.350\end{tabular} & \begin{tabular}[c]{@{}l@{}}0.350\\ 0.360\\ \textbf{0.252}\\ 0.322\\ 0.356\\ 0.344\end{tabular} & \begin{tabular}[c]{@{}l@{}}0.530\\ 0.588\\ 0.549\\ 0.515\\ \textbf{0.613}\\ \textbf{0.613}\end{tabular} & \begin{tabular}[c]{@{}l@{}}0.544\\ 0.516\\ 0.468\\ 0.544\\ 0.544\\ \textbf{0.550}\end{tabular} & \begin{tabular}[c]{@{}l@{}}0.251\\ 0.275\\ \textbf{0.115}\\ 0.129\\ 0.335\\ 0.346\end{tabular} & \begin{tabular}[c]{@{}l@{}}0.304\\ 0.360\\ \textbf{0.252}\\ 0.268\\ 0.370\\ 0.372\end{tabular} & \begin{tabular}[c]{@{}l@{}}0.562\\ \textbf{0.588}\\ 0.581\\ 0.540\\ 0.524\\ 0.536\end{tabular} \\ \hline
claude3-sonnet & \begin{tabular}[c]{@{}l@{}}Vanilla\\ CoT\\ Atypical scenario\\ Atypical presentations\\ Consistency (k=3)\\ Average (k=3)\end{tabular} & \begin{tabular}[c]{@{}l@{}}0.541\\ \textbf{0.638}\\ 0.564\\ 0.568\\ 0.552\\ 0.558\end{tabular} & \begin{tabular}[c]{@{}l@{}}0.331\\ 0.246\\ \textbf{0.124}\\ 0.136\\ 0.335\\ 0.338\end{tabular} & \begin{tabular}[c]{@{}l@{}}0.336\\ 0.282\\ \textbf{0.259}\\ 0.332\\ 0.363\\ 0.358\end{tabular} & \begin{tabular}[c]{@{}l@{}}\textbf{0.613}\\ 0.599\\ 0.547\\ 0.564\\ 0.555\\ 0.565\end{tabular} & \begin{tabular}[c]{@{}l@{}}0.565\\ \textbf{0.612}\\ 0.561\\ 0.531\\ 0.568\\ 0.568\end{tabular} & \begin{tabular}[c]{@{}l@{}}0.306\\ 0.265\\ \textbf{0.134}\\ 0.305\\ 0.346\\ 0.337\end{tabular} & \begin{tabular}[c]{@{}l@{}}0.327\\ 0.295\\ \textbf{0.268}\\ 0.316\\ 0.355\\ 0.356\end{tabular} & \begin{tabular}[c]{@{}l@{}}0.630\\ 0.615\\ 0.604\\ \textbf{0.666}\\ 0.591\\ 0.585\end{tabular} & \begin{tabular}[c]{@{}l@{}}0.128\\ \textbf{0.246}\\ 0.140\\ 0.100\\ 0.122\\ 0.128\end{tabular}  & \begin{tabular}[c]{@{}l@{}}0.569\\ 0.542\\ \textbf{0.438}\\ 0.517\\ 0.789\\ 0.750\end{tabular}  & \begin{tabular}[c]{@{}l@{}}0.428\\ 0.469\\ \textbf{0.252}\\ 0.339\\ 0.766\\ 0.725\end{tabular} & \begin{tabular}[c]{@{}l@{}}0.743\\ 0.663\\ 0.634\\ \textbf{0.880}\\ 0.443\\ 0.491\end{tabular} \\ \hline
gemini-pro-1.0 & \begin{tabular}[c]{@{}l@{}}Vanilla\\ CoT\\ Atypical scenario\\ Atypical presentations\\ Consistency (k=3)\\ Average (k=3)\end{tabular} & \begin{tabular}[c]{@{}l@{}}0.472\\ 0.465\\ 0.473\\ 0.458\\ 0.471\\ \textbf{0.477}\end{tabular} & \begin{tabular}[c]{@{}l@{}}0.369\\ 0.357\\ \textbf{0.105}\\ 0.293\\ 0.399\\ 0.364\end{tabular} & \begin{tabular}[c]{@{}l@{}}0.385\\ 0.369\\ \textbf{0.268}\\ 0.332\\ 0.400\\ 0.391\end{tabular} & \begin{tabular}[c]{@{}l@{}}0.530\\ 0.578\\ 0.510\\ 0.568\\ \textbf{0.613}\\ 0.599\end{tabular} & \begin{tabular}[c]{@{}l@{}}0.551\\ 0.526\\ 0.513\\ 0.387\\ \textbf{0.557}\\ 0.549\end{tabular} & \begin{tabular}[c]{@{}l@{}}0.297\\ 0.306\\ \textbf{0.129}\\ 0.357\\ 0.337\\ 0.314\end{tabular} & \begin{tabular}[c]{@{}l@{}}0.338\\ 0.340\\ \textbf{0.274}\\ 0.316\\ 0.343\\ 0.341\end{tabular} & \begin{tabular}[c]{@{}l@{}}0.520\\ 0.537\\ 0.517\\ \textbf{0.712}\\ 0.624\\ 0.635\end{tabular} & \begin{tabular}[c]{@{}l@{}}0.492\\ 0.438\\ 0.508\\ 0.226\\ \textbf{0.540}\\ 0.504\end{tabular} & \begin{tabular}[c]{@{}l@{}}0.342\\ 0.368\\ \textbf{0.128}\\ 0.448\\ 0.309\\ 0.325\end{tabular}  & \begin{tabular}[c]{@{}l@{}}0.362\\ 0.373\\ \textbf{0.276}\\ 0.338\\ 0.349\\ 0.371\end{tabular} & \begin{tabular}[c]{@{}l@{}}0.572\\ 0.590\\ 0.495\\ \textbf{0.782}\\ 0.591\\ 0.529\end{tabular} \\ \hline
gpt-4-turbo    & \begin{tabular}[c]{@{}l@{}}Vanilla\\ CoT\\ Atypical scenario\\ Atypical presentations\\ Consistency (k=3)\\ Average (k=3)\end{tabular} & \begin{tabular}[c]{@{}l@{}}0.756\\ \textbf{0.832}\\ 0.741\\ 0.751\\ 0.775\\ 0.767\end{tabular} & \begin{tabular}[c]{@{}l@{}}0.133\\ \textbf{0.065}\\ 0.085\\ 0.114\\ 0.198\\ 0.194\end{tabular} & \begin{tabular}[c]{@{}l@{}}0.190\\ \textbf{0.132}\\ 0.181\\ 0.178\\ 0.206\\ 0.205\end{tabular} & \begin{tabular}[c]{@{}l@{}}0.670\\ \textbf{0.710}\\ 0.693\\ 0.673\\ 0.555\\ 0.573\end{tabular} & \begin{tabular}[c]{@{}l@{}}0.707\\ \textbf{0.730}\\ 0.675\\ 0.681\\ 0.712\\ 0.708\end{tabular} & \begin{tabular}[c]{@{}l@{}}0.188\\ 0.162\\ \textbf{0.071}\\ 0.174\\ 0.248\\ 0.249\end{tabular} & \begin{tabular}[c]{@{}l@{}}0.230\\ \textbf{0.206}\\ \textbf{0.206}\\ 0.213\\ 0.253\\ 0.255\end{tabular} & \begin{tabular}[c]{@{}l@{}}0.673\\ 0.729\\ 0.672\\ \textbf{0.739}\\ 0.587\\ 0.590\end{tabular} & \begin{tabular}[c]{@{}l@{}}0.394\\ 0.358\\ 0.354\\ 0.338\\ \textbf{0.404}\\ \textbf{0.404}\end{tabular} & \begin{tabular}[c]{@{}l@{}}0.374\\ 0.445\\ \textbf{0.197}\\ 0.414\\ 0.537\\ 0.552\end{tabular}  & \begin{tabular}[c]{@{}l@{}}0.337\\ 0.402\\ \textbf{0.249}\\ 0.363\\ 0.546\\ 0.563\end{tabular} & \begin{tabular}[c]{@{}l@{}}\textbf{0.792}\\ 0.743\\ 0.679\\ 0.763\\ 0.490\\ 0.458\end{tabular} \\ \hline
\end{tabular}
}
\caption{\label{tab:results}
    Using atypicality as post-hoc calibration brings major improvements in ECE and Brier Scores across all datasets and all models. \textbf{Atypical Scenario} outperforms all other methods in calibration in the big majority of experiments.
  }
\end{table*}

\paragraph{Evaluation Metrics}
To measure how well the confidence estimates are calibrated, we will report multiple metrics across the different datasets, methods and models. Calibration is defined as how well a model's predicted probability is aligned with the true likelihoods of outcomes \citep{yuksekgonul2023confidence, doi:10.1198/016214506000001437}. We measure this using \textit{Expected Calibration Error (ECE)} \citep{naeini2015obtaining} and \textit{Brier Score} \citep{1180014634}. 

To evaluate the quality of confidence estimates using ECE, we group the model's confidence into $K$ bins and estimate ECE by taking the weighted average of the difference between confidence and accuracy in each bin \citep{he2023investigating}. Formally, let $N$ be the sample size, $K$ the number of bins, and $I_k$ the indices of samples in the $k^{th}$ bin, we have:
\begin{equation}
    ECE_K = \sum_{k=1}^K \frac{|I_k|}{N} |acc(I_k) - conf(I_k)|
\end{equation}

Brier score is a scoring function that measures the accuracy of the predicted confidence estimates and is equivalent to the mean squared error. Formally, it is defined as:
\begin{equation}
    BS = \frac{1}{N} \sum_{n=1}^N (conf_n - o_n)^2
\end{equation}
where $conf_n$ and $o_n$ are the confidence estimate and outcome of the $n^{th}$ sample respectively. 

Additionally, to evaluate if the LLM can convey higher confidence scores for correct predictions and lower confidence scores for incorrect predictions, we use the \textit{Area Under the Receiver Operating Characteristic Curve (AUROC)}. Finally, to assess any significant changes in performance, we also report \textit{accuracy} on the different tasks.

\subsection{Results and Analysis}
\begin{figure}[ht]
  \includegraphics[width=1\columnwidth]{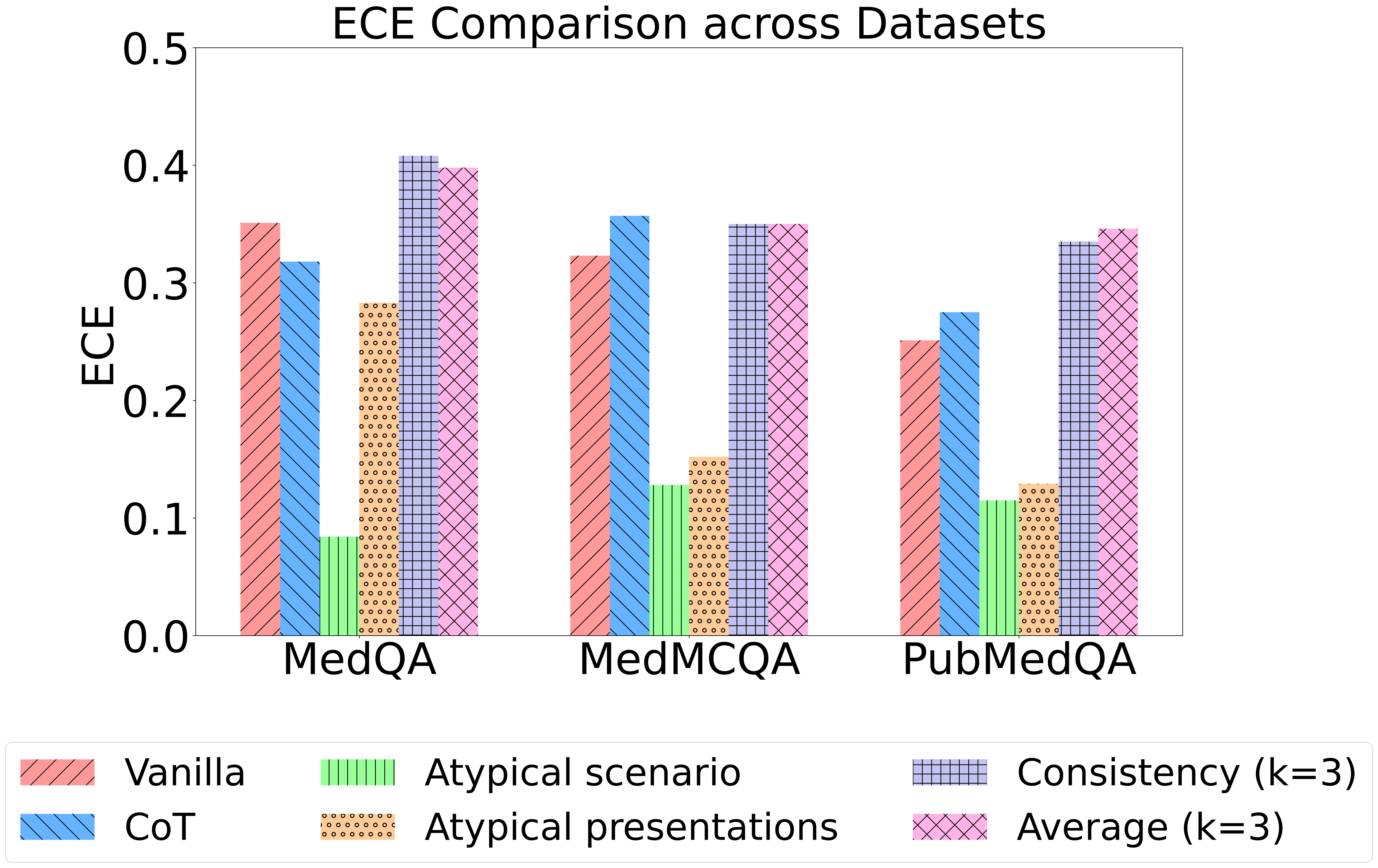} \hfill
  \caption{ECE of GPT-3.5-turbo for each method across all three datasets.}
  \label{fig:ece_per_dataset}
\end{figure}

To assess the ability of LLMs to provide calibrated confidence scores and explore the use of atypical scores for calibration, we experimented with each mentioned method using four different black-box LLMs across three medical question-answering datasets. The main results and findings are reported in the following section.

\begin{figure*}[t]
  \includegraphics[width=0.48\linewidth]{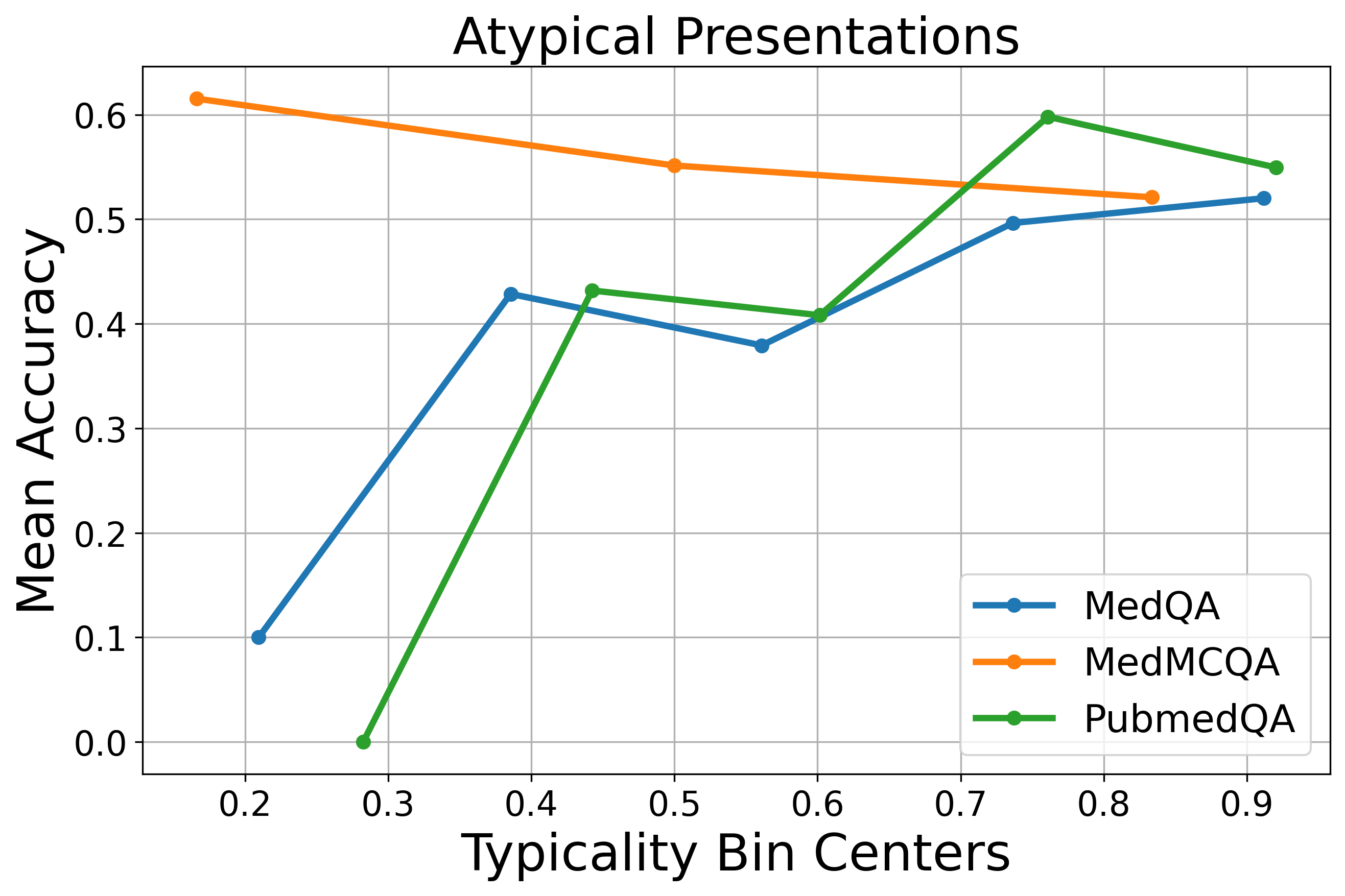} \hfill
  \includegraphics[width=0.48\linewidth]{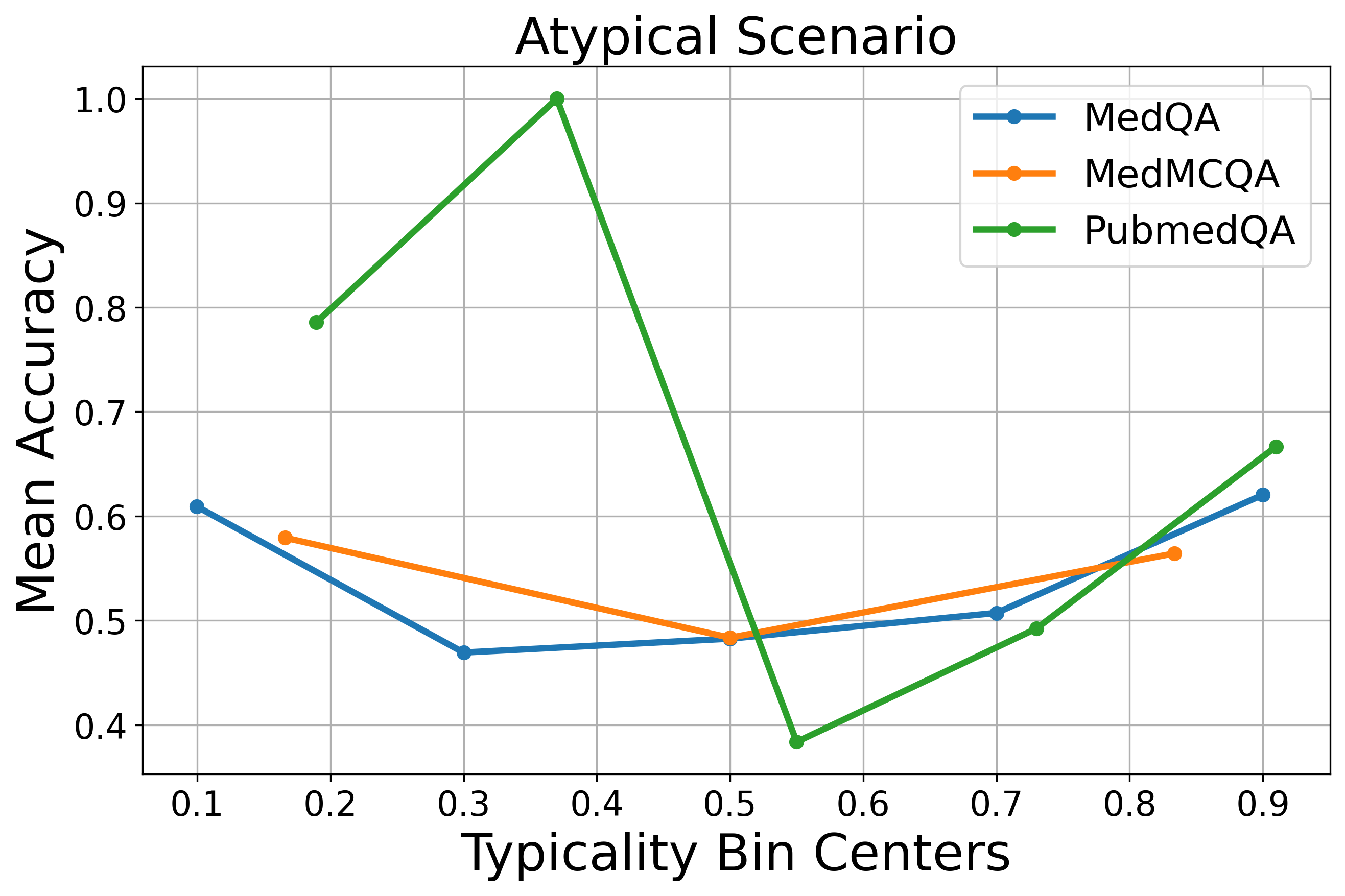}
  \caption {Accuracy by Typicality bins of GPT3.5-turbo for Atypical Presentations Aware Recalibration methods.}
  \label{fig:atypicality_bins_behaviour}
\end{figure*}

\paragraph{LLMs are miscalibrated for Medical QA.}
To evaluate the reliability and calibration of confidence scores elicited by the LLMs, we examined the calibration curves of GPT-3.5-turbo in Figure \ref{fig:calibration_curves}, where the green dotted line represents perfect calibration. The results indicate that the confidence scores are generally miscalibrated, with the LLMs tending to be overconfident. Although the \textit{Atypical Scenario} and \textit{Atypical Presentations} methods show improvements with better alignment, there is still room for improvement. Introducing recalibration methods with atypicality scores results in more variation in the calibration curves, including instances of underestimation. Additional calibration curves for the other models are provided in Appendix \ref{sec:appendixA}.

\paragraph{Leveraging atypical scores greatly improves calibration.}
We analyzed the calibration metrics for each method and found that leveraging atypical scores significantly reduces ECE and Brier Score across all datasets, as shown in Figure \ref{fig:ece_per_dataset} and Table \ref{tab:results}. In contrast, other methods show minor changes in calibration errors, with some even increasing ECE. The \textit{Consistency} and \textit{Average} methods do not show improvement, and sometimes degrade, due to the multiple-choice format of the datasets, which shifts confidence estimates to higher, more overconfident values. However, the \textit{Atypical Scenario} method, which elicits an atypical score describing how unusual the medical scenario is, outperforms all other methods and significantly lowers ECE compared to vanilla confidence scores. It is very interesting that the level of atypicality considered seems to make a significant difference. It is a hallmark of reasoning that how the LLM aggregates the atypicality from a lower level when prompted for a scenario is superior to simply aggregating the symptoms atypicality. This opens for further investigation into how LLMs reason about atypicality. We discuss and analyze the role of atypicality in calibration in the following sections. Detailed results of our experiments are reported in Table \ref{tab:results}.

\begin{figure}[ht]
  \includegraphics[width=1\columnwidth]{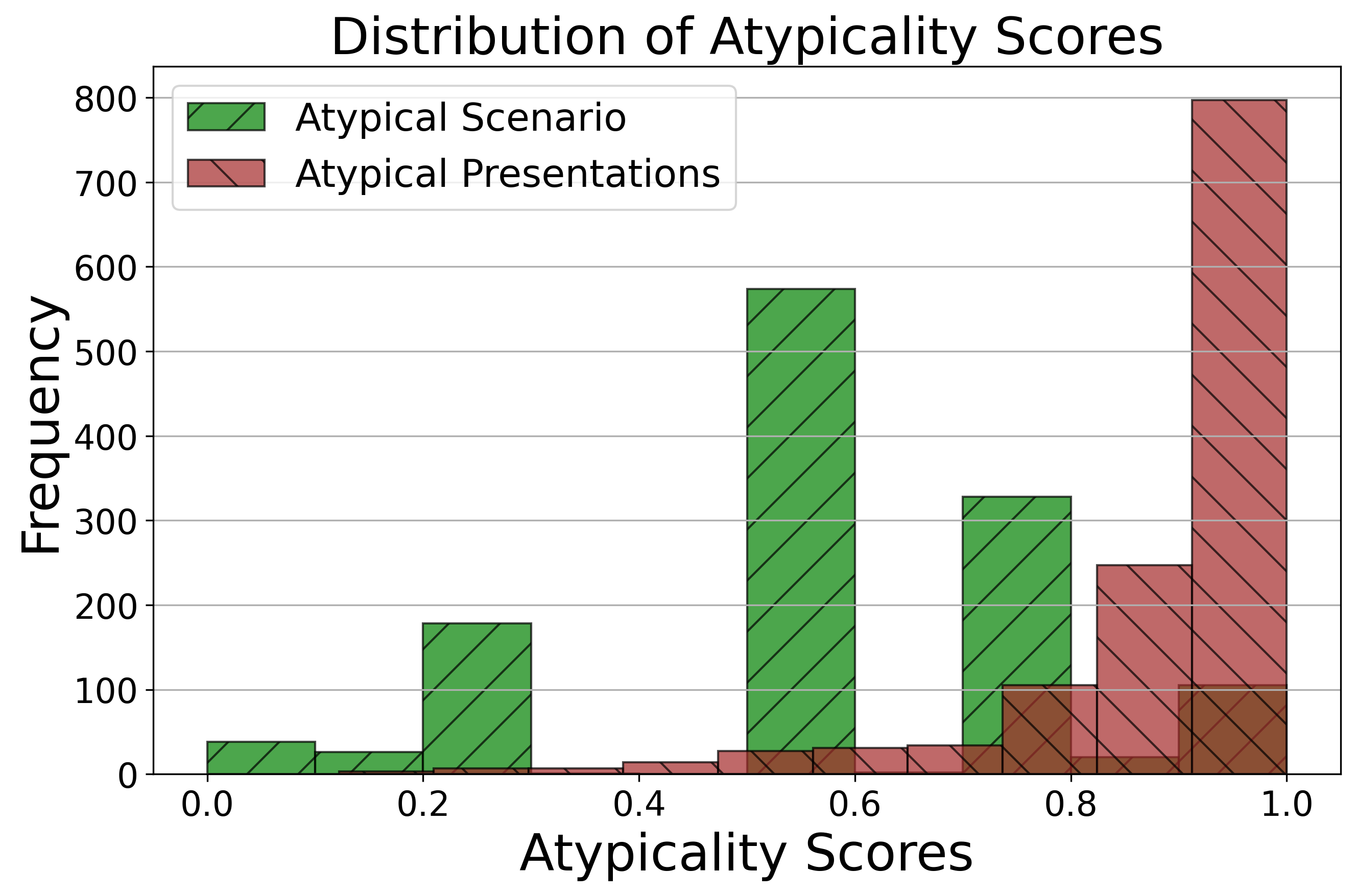} \hfill
  \caption{Distribution of atypicality scores between Atypical Presentations and Atypical Scenario of GPT-3.5-turbo on MedQA.}
  \label{fig:atypicality_distributions}
\end{figure}
% \begin{figure*}[t]
%   \includegraphics[width=0.48\linewidth]{images/gpt3p5/acc_vs_atypicality.png} \hfill
%   \includegraphics[width=0.48\linewidth]{images/gpt3p5/ece_vs_atypicality.png}
%   \caption {Performance and Calibration by Atypicality bins for gpt-3.5-turbo on MedQA}
%   \label{fig:atypicality_bins_behaviour}
% \end{figure*}
\paragraph{Atypicality distribution varies between Atypical Scenario and Atypical Presentations.}
% To better understand the gap between the calibration errors of \textbf{Atypical Scenario} and \textit{Atypical Presentations}, we first look into the distribution of the atypicality scores. In Figure \ref{fig:atypicality_distributions}, we see that the distribution of \textit{Atypical Presentations} is much more right skewed meaning that there is prevalence of typical scores. This is greatly due to the nature of the approach. As not all questions in the datasets are necessarily diagnostic questions, for example questions asking for medical advice, there is no atypicality associated with symptoms or presentations. In our framework, we simply impute the atypicality score to 1 so that it doesn't affect the original confidence estimate. In contrast, \textit{Atypical Scenario} shows a more distributed spread over the scores. This suggests that the LLMs are able to identify that some questions and scenarios are more atypical, which allows to take that into account when calibrating the confidence estimates.
To better understand the gap between the calibration errors of \textit{Atypical Scenario} and \textit{Atypical Presentations}, we first examine the distribution of the atypicality scores. In Figure \ref{fig:atypicality_distributions}, we observe that the distribution of \textit{Atypical Presentations} is much more right-skewed, indicating a prevalence of typical scores. This is largely due to the nature of the approach. Not all questions in the datasets are necessarily diagnostic questions; for example, some may ask for medical advice, where there is no atypicality associated with symptoms or presentations. In our framework, we impute the atypicality score to 1 for such cases, so it does not affect the original confidence estimate. In contrast, \textit{Atypical Scenario} shows a more evenly distributed spread over the scores. This suggests that the LLMs can identify that some questions and scenarios are more atypical, which allows this atypicality to be considered when calibrating the confidence estimates.

\begin{figure*}[t]
  \includegraphics[width=0.48\linewidth]{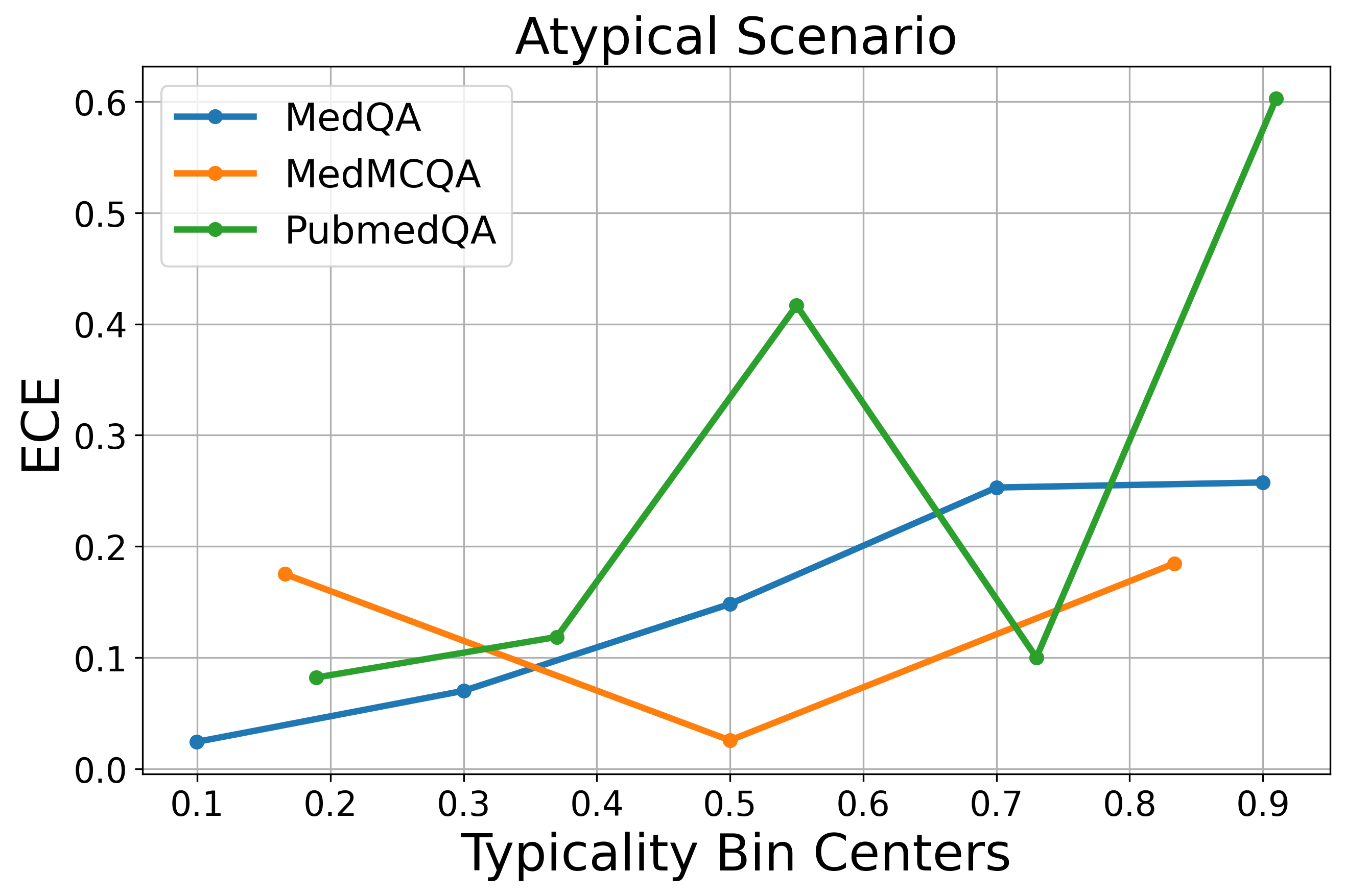} \hfill
  \includegraphics[width=0.48\linewidth]{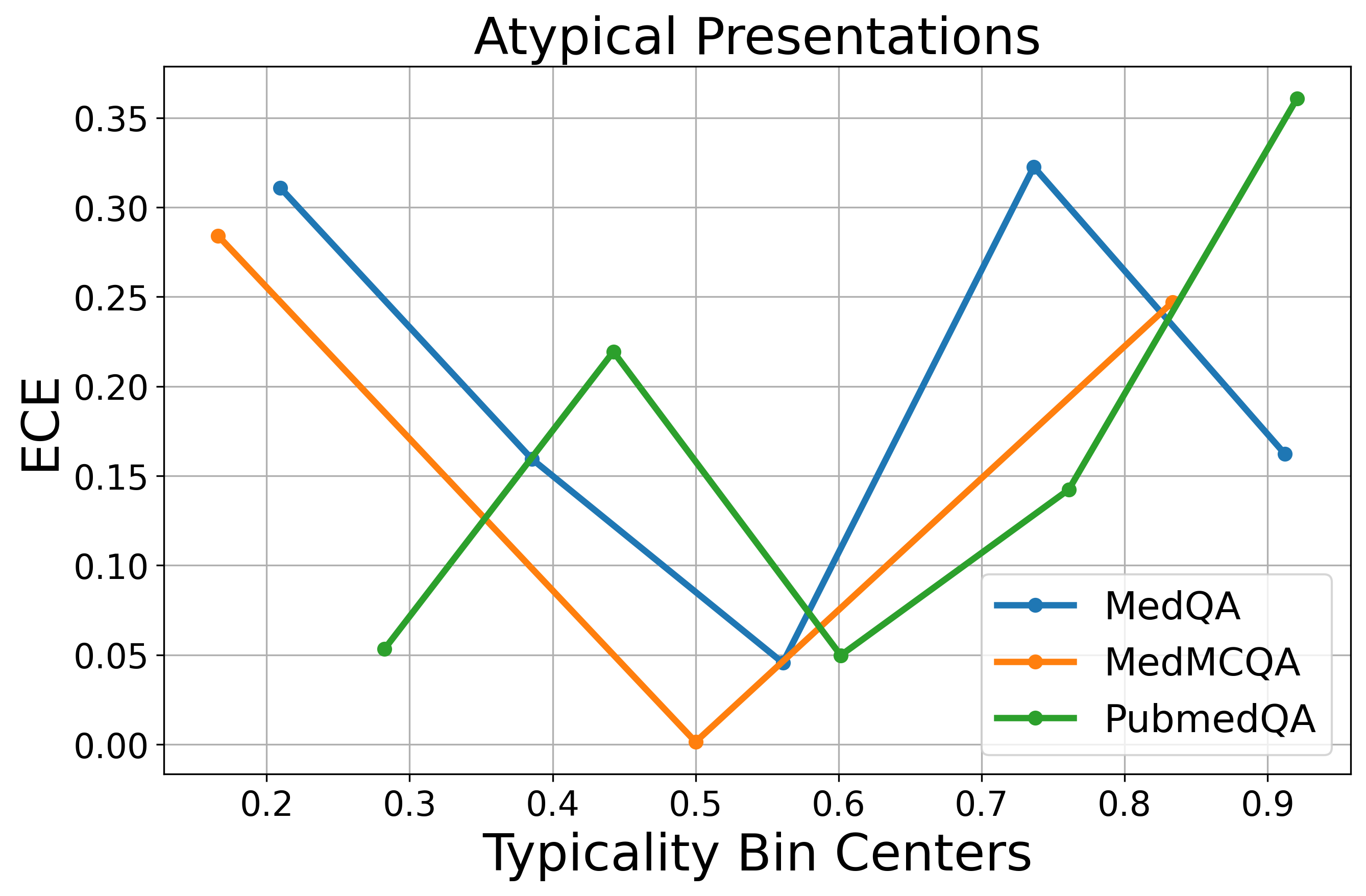}
  \caption {ECE by Typicality bins of GPT3.5-turbo for Atypical Presentations Aware Recalibration methods.}
  \label{fig:ece_by_bins_gpt3}
\end{figure*}

\paragraph{Typical samples do not consistently outperform atypical samples.}
We now question the performance of atypical versus typical samples. The intuitive answer is that performance should be better on typical samples, which are common scenarios or symptoms, making the question easier to answer. However, as shown in Figure \ref{fig:atypicality_bins_behaviour}, there is no consistent pattern between accuracy and atypicality for GPT-3.5-turbo. While accuracy increases as atypicality decreases in some cases like MedQA and PubMedQA, in other cases, the accuracy remains unchanged or even decreases. This performance variation across typicality bins provides insights into how LLMs use the notion of atypicality in their reasoning process. Higher accuracy for atypical samples could suggest that unique, easily identifiable features help the LLM. Conversely, high atypicality can indicate that the question is more difficult, leading to lower accuracy. To understand this better, we also experimented with prompts to retrieve difficulty scores and analyzed their relationship with atypicality. Our results show no clear correlation between difficulty and atypicality scores. Most atypicality scores are relatively high across all difficulty levels. Although some atypical samples are deemed more difficult, the results are inconsistent and hard to interpret. Associated graphs are in Appendix \ref{sec:appendixA}. Briefly, this inconsistent performance behavior shows there is more to explore about how LLMs use atypicality intrinsically.

\paragraph{Atypicality does not predict LLM's calibration error.}
Another question we explored was whether calibration errors correlate with atypicality. We used the same approach as our performance analysis, binning the samples by atypicality scores and examining the ECE within each bin. This allowed us to evaluate how well the model's predicted confidence level aligned with actual outcomes across varying levels of atypicality. For both \textit{Atypical Scenario} and \textit{Atypical Presentations}, we assessed GPT-3.5-turbo's calibration. As shown in Figure \ref{fig:atypicality_bins_behaviour}, there are no clear patterns between atypicality scores and calibration errors. The high fluctuation of ECE across different levels of atypicality suggests that the model experiences high calibration errors for both typical and atypical samples. This indicates that calibration performance is influenced by factors beyond just atypicality. Similar to the previous performance analysis in terms of accuracy, how LLMs interpret and leverage atypicality may vary between samples, leading to inconsistent behavior.

\paragraph{Atypicality helps in failure prediction.}
While ECE and Brier Score provide insights into the reliability and calibration of confidence estimates, it is also important for the model to assign higher confidences to correct predictions and lower confidences to incorrect predictions. To assess this, we used AUROC. In Table \ref{tab:results}, we observe that incorporating atypicality into our model improves its performance across most experiments compared to the vanilla baseline. However, these improvements do not consistently outperform all other methods evaluated. This indicates that, while incorporating atypicality can improve the model's failure prediction, there remain specific scenarios where alternative methods may be more effective.

% we see that in most experiments, the best method leverages atypicality scores. However, even in experiments where it is not the best for failure prediction, it still shows improvements compared to the vanilla method in most cases.

\section{Conclusion}
In our study, we have demonstrated that LLMs remain miscalibrated and overconfident in the medical domain. Our results indicate that incorporating the notion of atypicality when eliciting LLM confidence leads to significant gains in calibration and some improvement in failure prediction for medical QA tasks. This finding opens the door to further investigate the calibration of LLMs in other high-stakes domains. Additionally, it motivates the development of methods that leverage important domain-specific notions and adapting our method for white-box LLMs. We hope that our work can inspire others to tackle these challenges and to develop methods for more trustworthy, explainable and transparent models, which are becoming increasingly urgent.

% While our framework is not limited to black-box LLMs, white-box LLMs allow for more flexibility in incorporating and extracting atypicality or other relevant notions.

\paragraph{Limitations} This study present a first effort into assessing black-box LLMs calibration and the use of atypicality in the healthcare domain. Several aspects of the study can further be improved for a better assessment. While we restricted ourselves to three modical question-answering datasets, we can expand it to more datasets with questions that are more open-ended or even different tasks such as clinical notes summarization which could also benefit a lot from having trusted confidence estimates. Next, we limited to use only commercial LLMs as they sit better in the medical context because of their ease of use and availability. Since our approach is also applicable to open-source LLMs, testing and assessing our approach to these other models will allow for a more comprehensive review of calibration in LLMs for the medical setting. Morever, our approach is still dependent on a prompt, and since LLMs are quite sensitive to how we prompt them, there could be even more optimal prompts for retrieving atypicality scores. Lastly, the notion of atypicality is not only seen and leverage in healthcare, but it is also present in other domains such as law. Adapting our methodology for other domains could further improve LLMs calibration performance.

\paragraph{Ethical considerations} In our work, we focus on the medical domain with the goal of enhancing the calibration and accuracy of confidence scores provided by large language models to support better-informed decision-making. While our results demonstrate significant improvements in calibration, it is imperative to stress that LLMs should not be solely relied upon without the oversight of a qualified medical expert. The involvement of a physician or an expert is essential to validate the model's recommandations and ensure a safe and effective decision-making process.

Moreover, we acknowledge the ethical implications of deploying AI in healthcare. It is crucial to recognize that LLMs are not infallible and can produce erroneuous outputs. Ensuring transparency in how these models reach their conclusions, and incorporating feedback from healthcare professionals are vital steps in maintaining the integrity and safety of medical practice. Thus, our work is a step towards creating reliable tools, but it must be integrated thoughtfully within the existing healthcare framework to truly benefit patient outcomes.
\bibliography{custom}

\appendix

\section{Additional Results}
\begin{figure}[ht]
  \includegraphics[width=1\columnwidth]{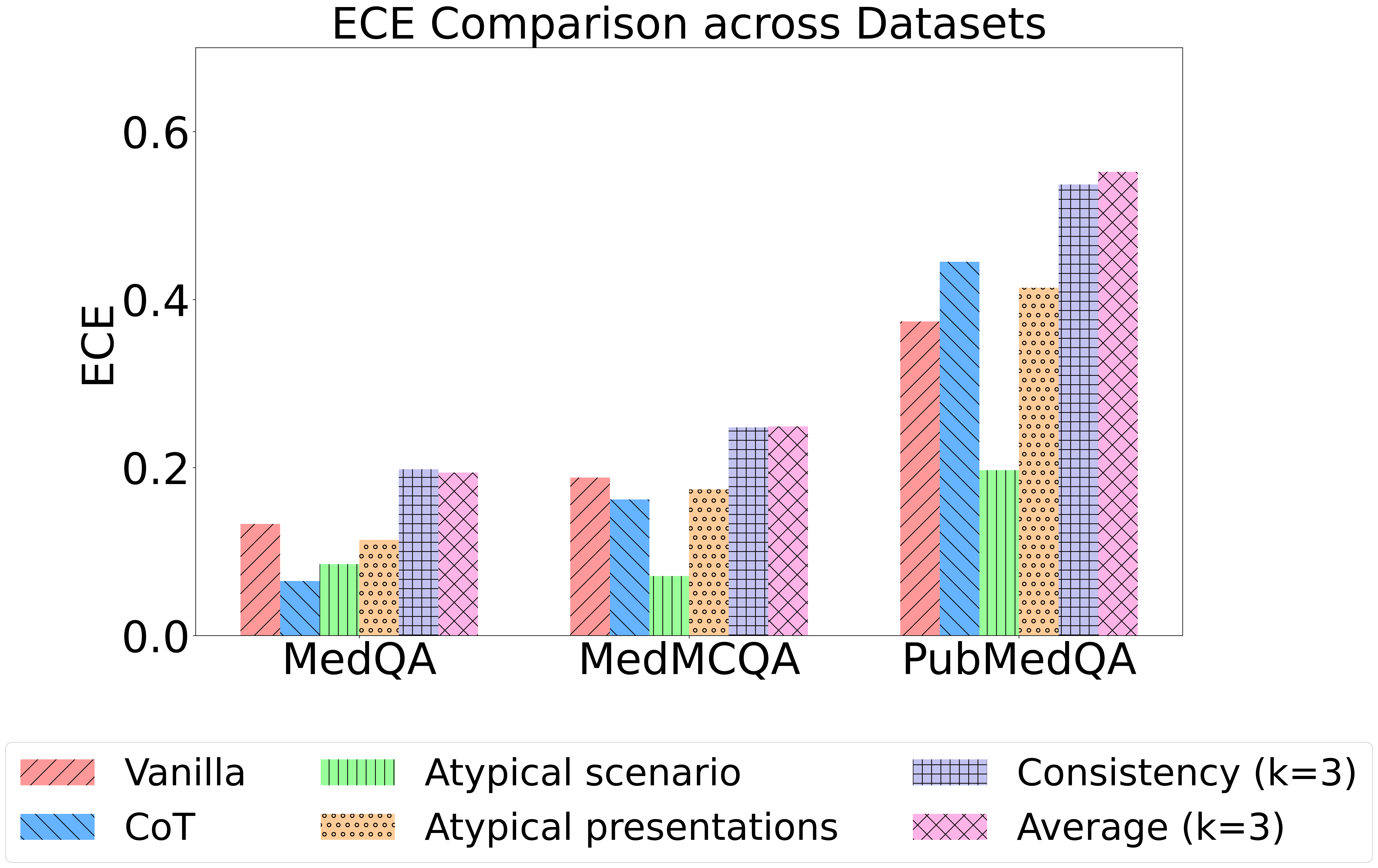} \hfill
  \caption{GPT4 ECE comparison across datasets}
  \label{fig:ece_per_dataset_gpt4}
\end{figure}
\begin{figure}[ht]
  \includegraphics[width=1\columnwidth]{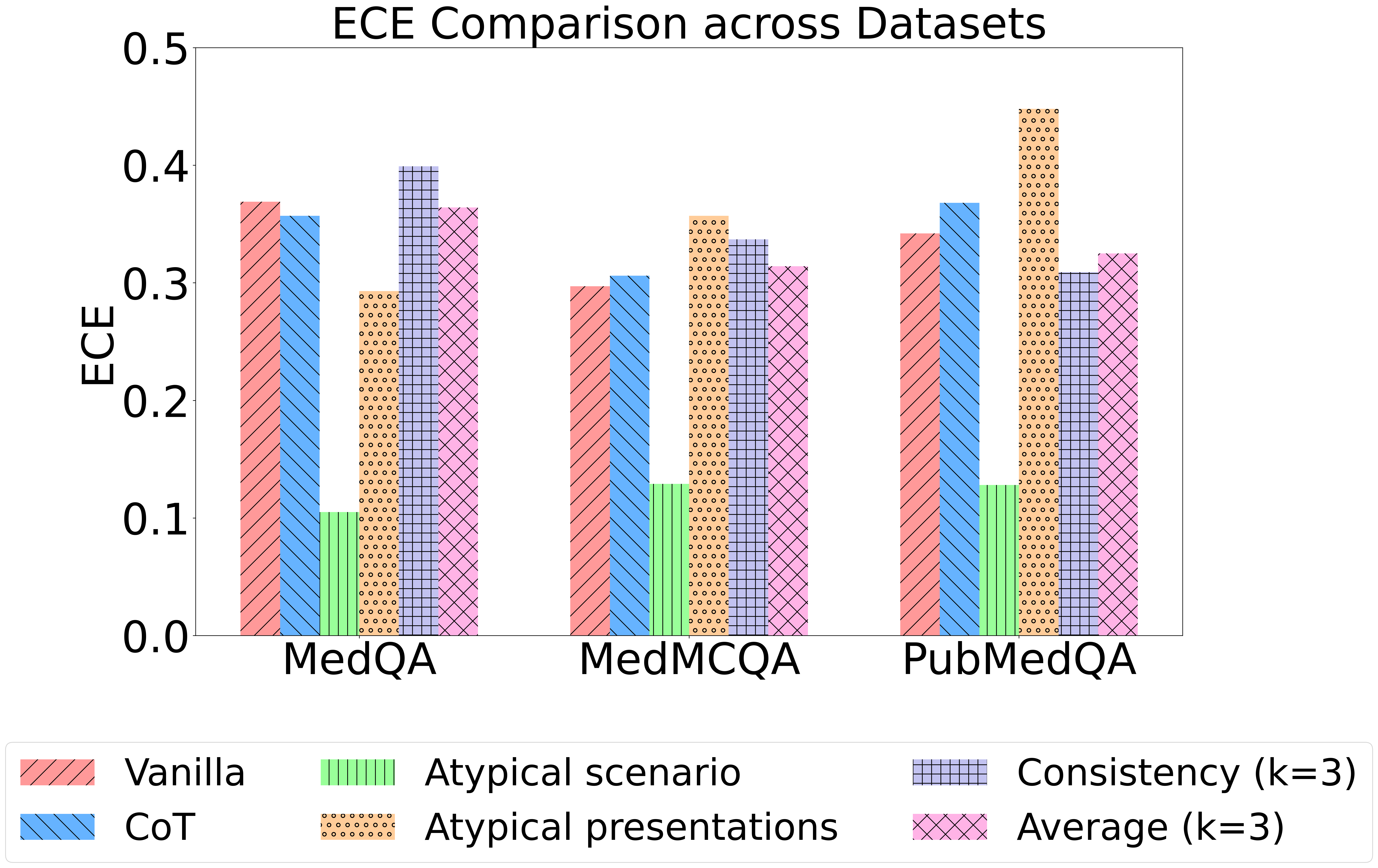} \hfill
  \caption{Gemini ECE comparison across datasets}
  \label{fig:ece_per_dataset_gemini}
\end{figure}
\begin{figure}[ht]
  \includegraphics[width=1\columnwidth]{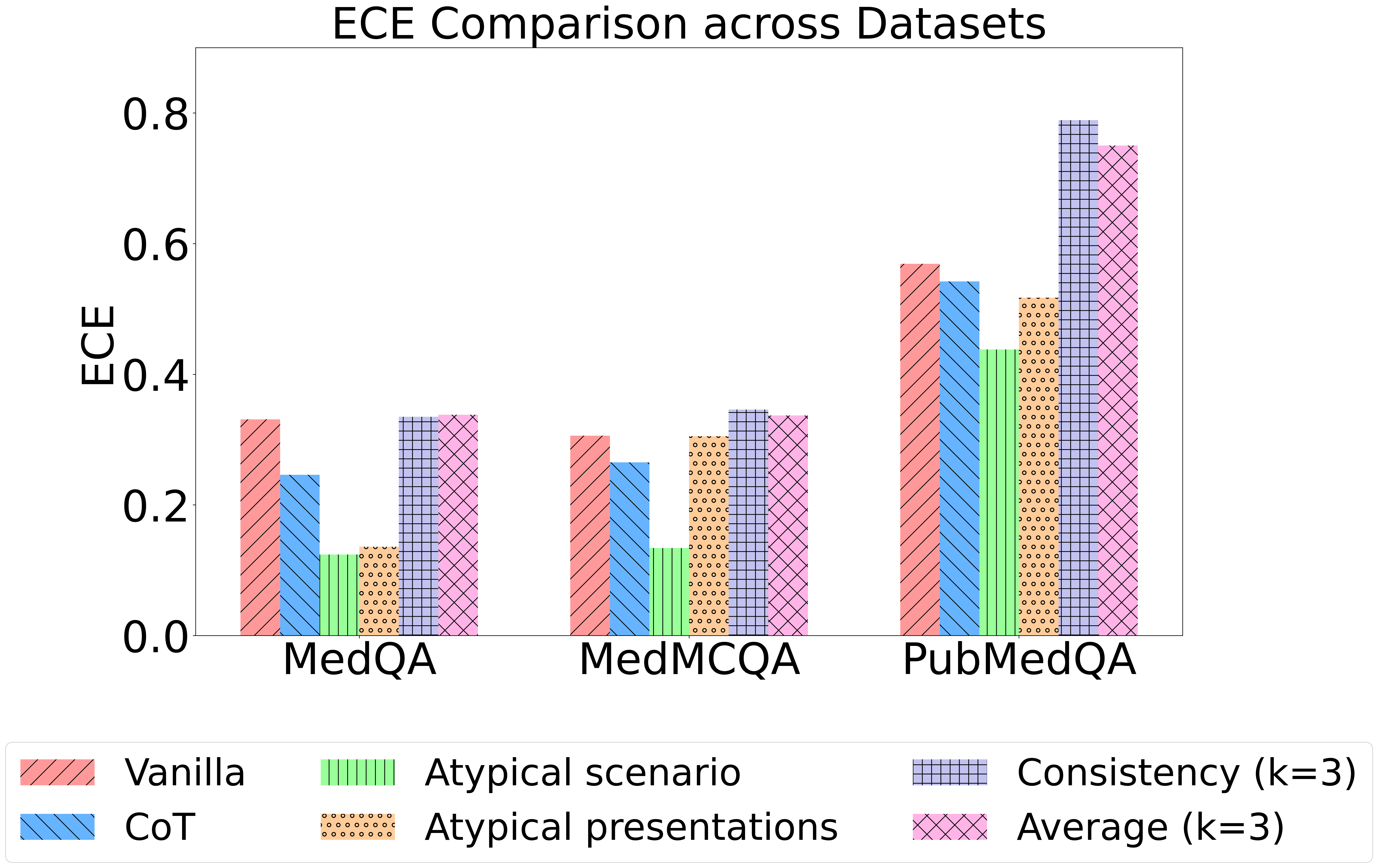} \hfill
  \caption{Claude3-sonnet ECE comparison across datasets}
  \label{fig:ece_per_dataset_claude}
\end{figure}

\label{sec:appendixA}
In the main sections of the paper, we presented figures for GPT3.5-turbo. Here we provide additional results for GPT3.5-turbo and the other three models to support the claims and findings discussed above. We show calibration and performance metrics for all methods used and across all three datasets: MedQA, MedMCQA and PubmedQA. Furthermore, we provide additional graphs to support the analysis of the distributions of atypicality scores across the different datasets as well as the distribution of atypicality scores by difficulty levels.

The findings and conclusions from these additional figures are already discussed in the main sections of the paper. These supplementary figures are included here to demonstrate that the findings are consistent across multiple models, ensuring that the conclusions drawn are robust and not based solely on one model.

\begin{figure*}[ht]
  \includegraphics[width=0.48\linewidth]{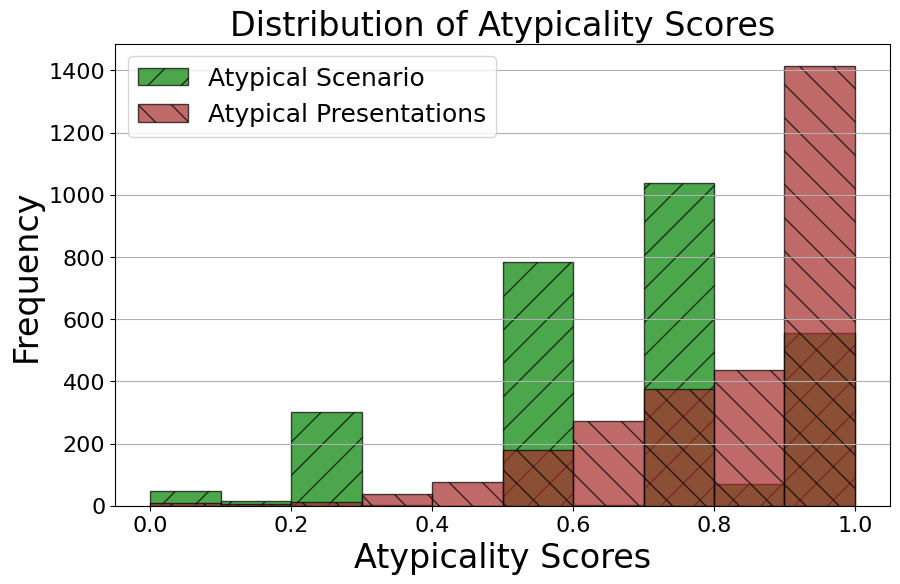}
  \includegraphics[width=0.48\linewidth]{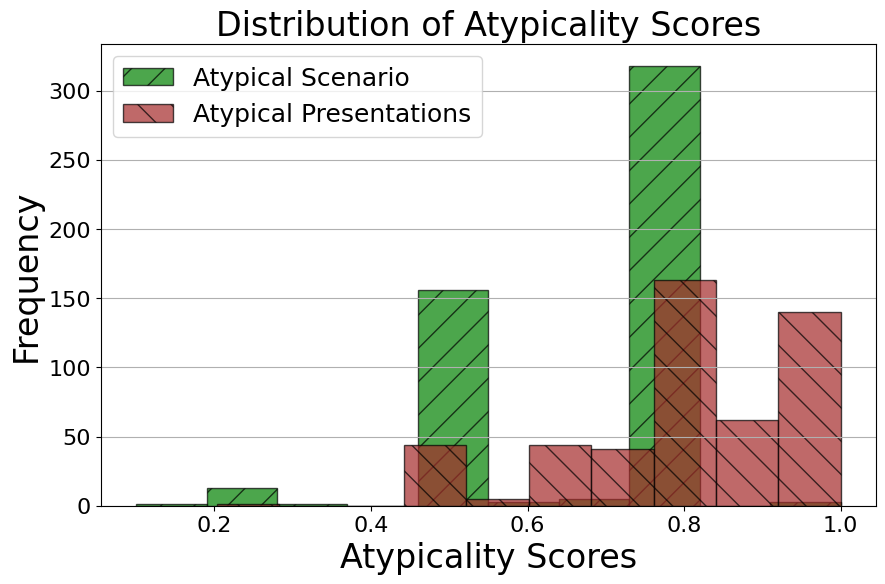}
  \caption {Atypicality Distributions of GPT3.5}
  \label{fig:atypicality_dist_gpt3}
\end{figure*}

\begin{figure*}[ht]
  \includegraphics[width=0.32\linewidth]{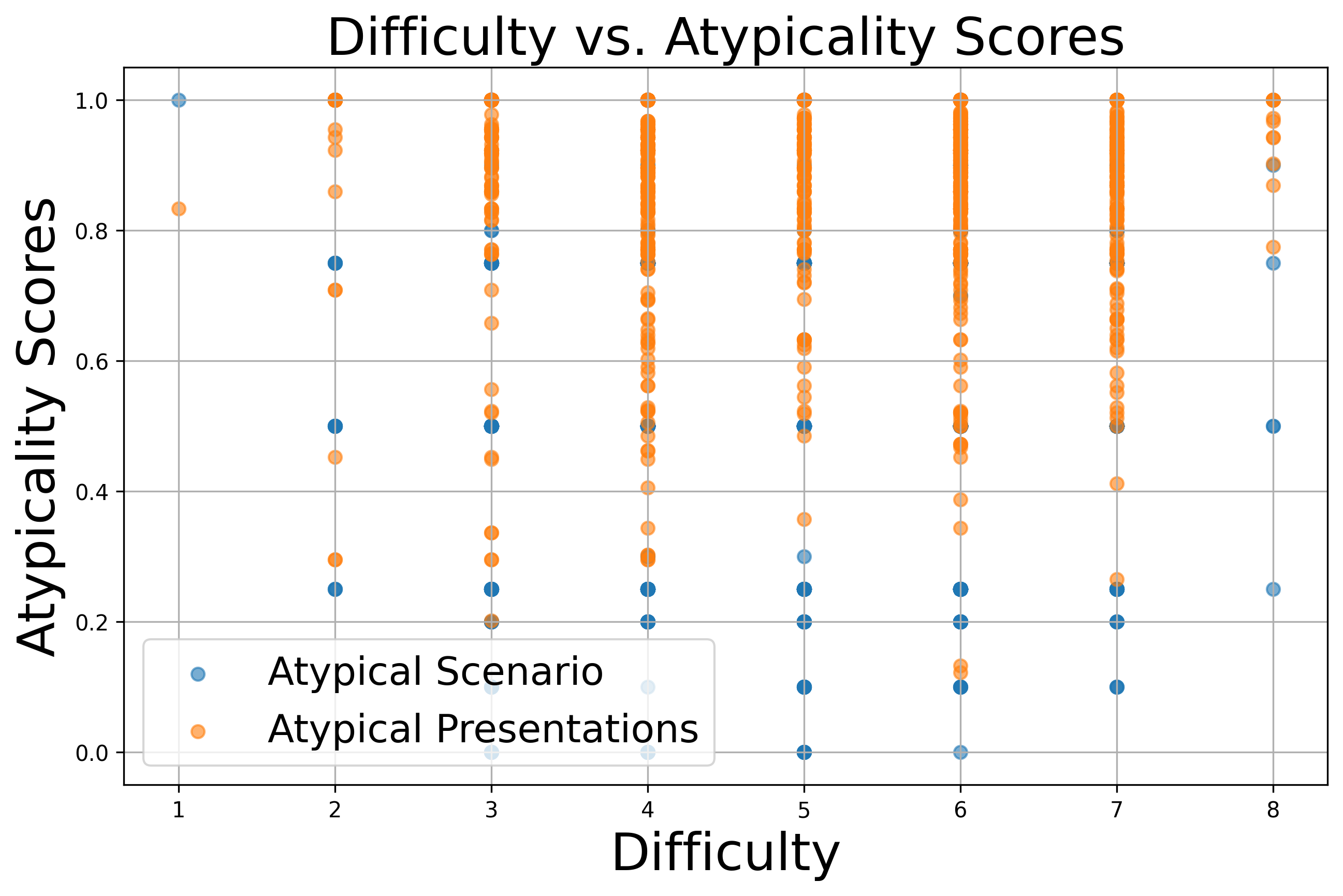} \hfill
  \includegraphics[width=0.32\linewidth]{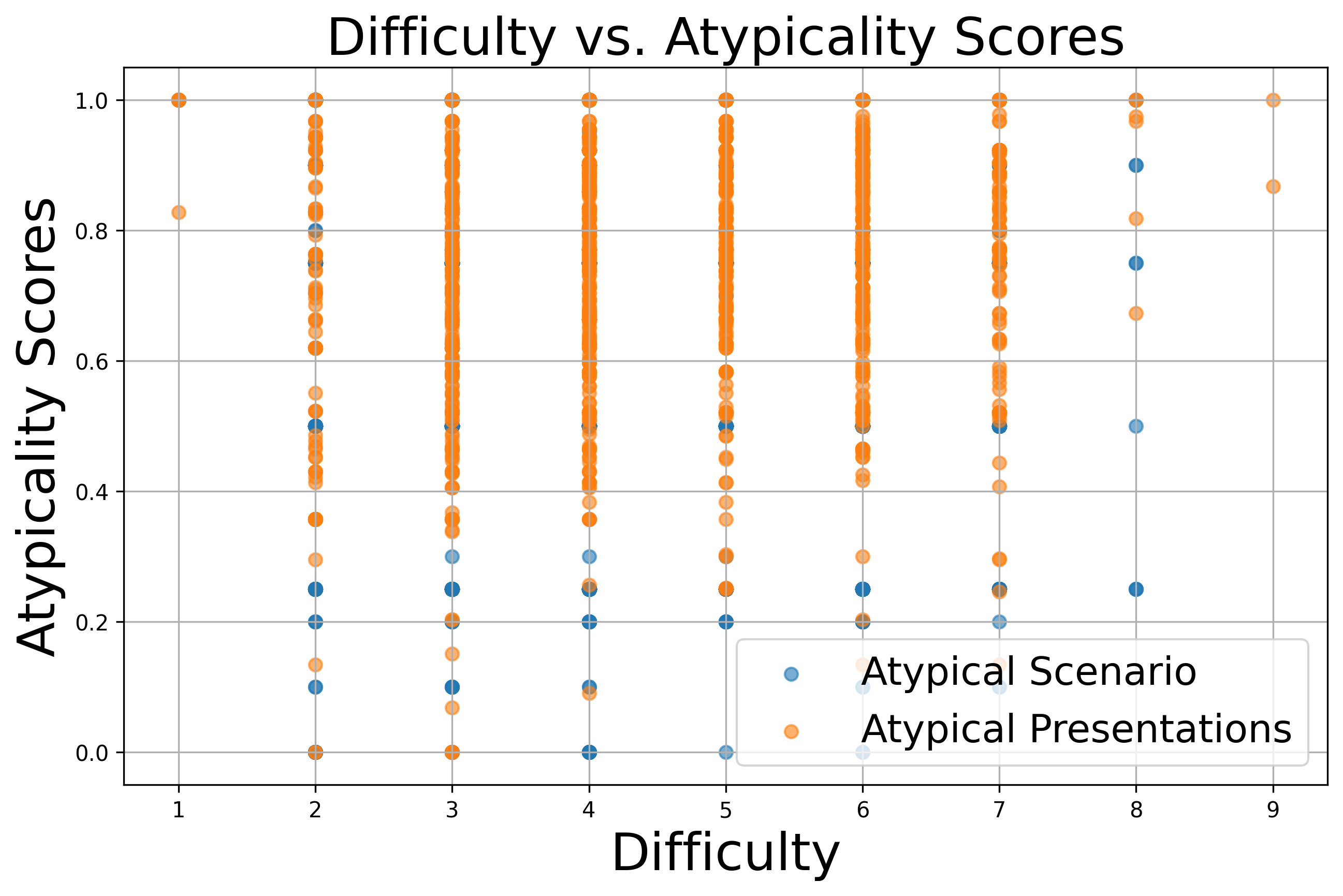}
  \includegraphics[width=0.32\linewidth]{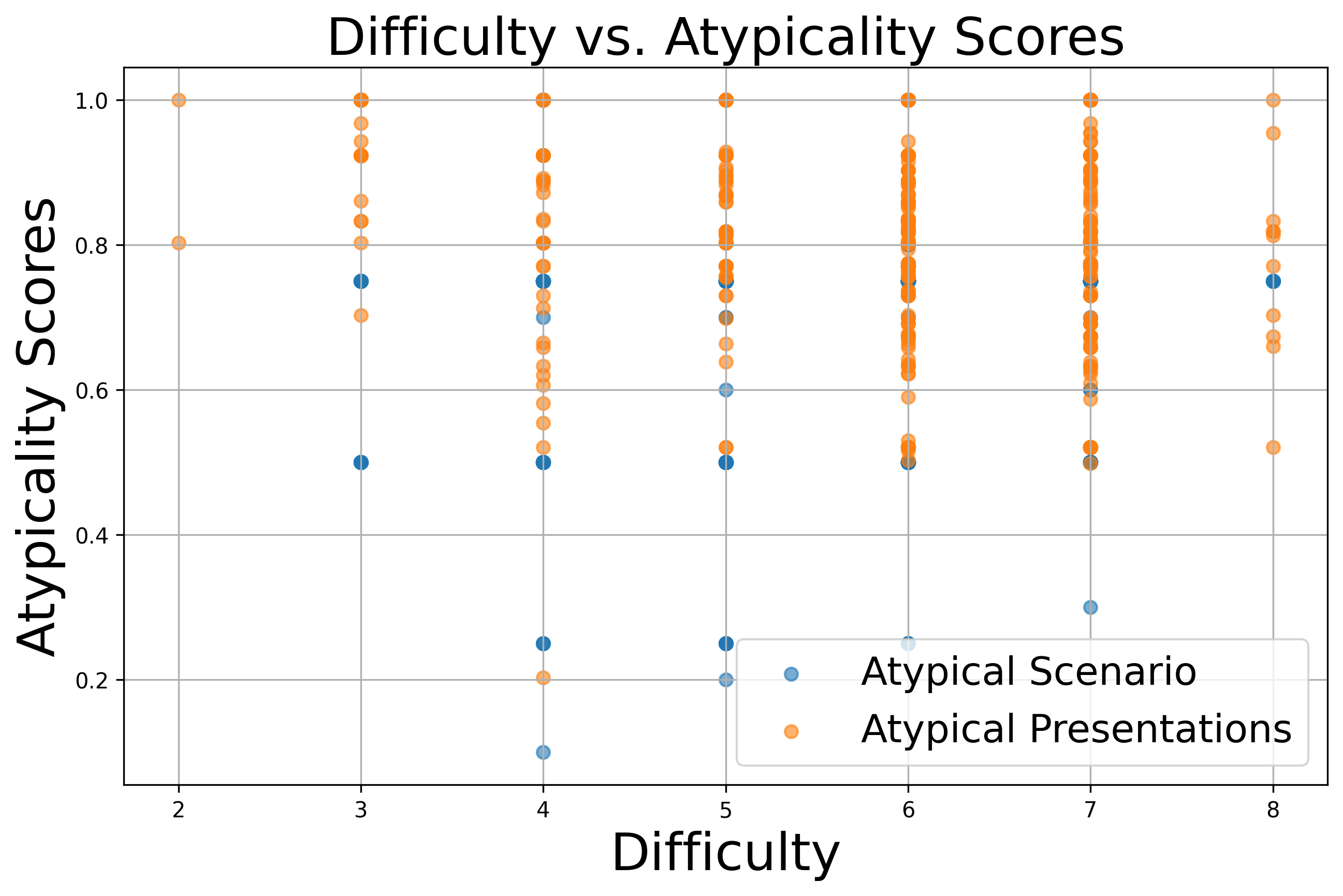}
  \caption {Atypicality by Difficulty for GPT3.5}
  \label{fig:diff_dist_gpt3}
\end{figure*}

\begin{figure*}[ht]
  \includegraphics[width=1\linewidth]{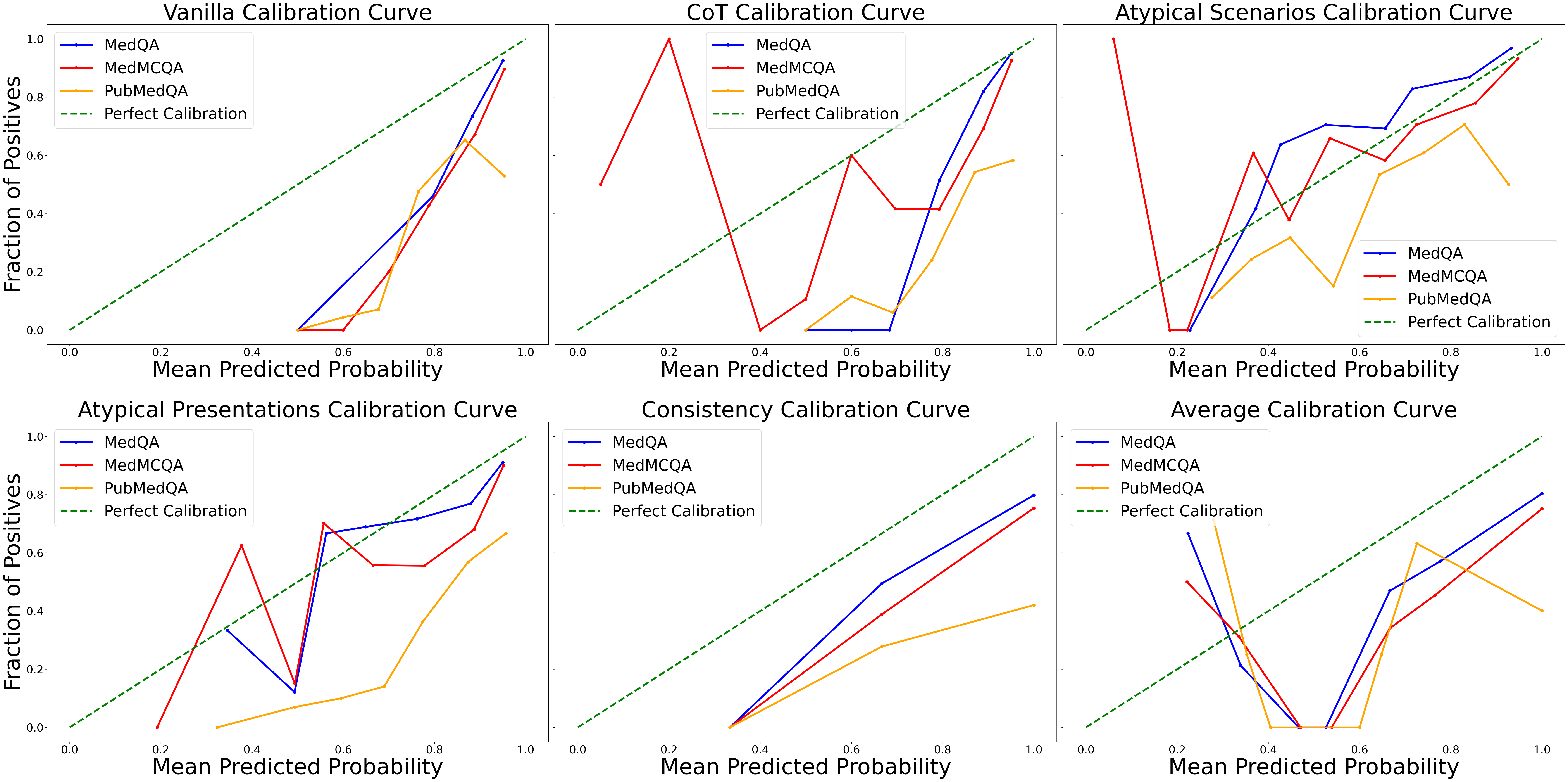} \hfill
  \caption {GPT4 Calibration curves across all methods}
  \label{fig:calibration_curves_gpt4}
\end{figure*}

\begin{figure*}[ht]
  \includegraphics[width=0.48\linewidth]{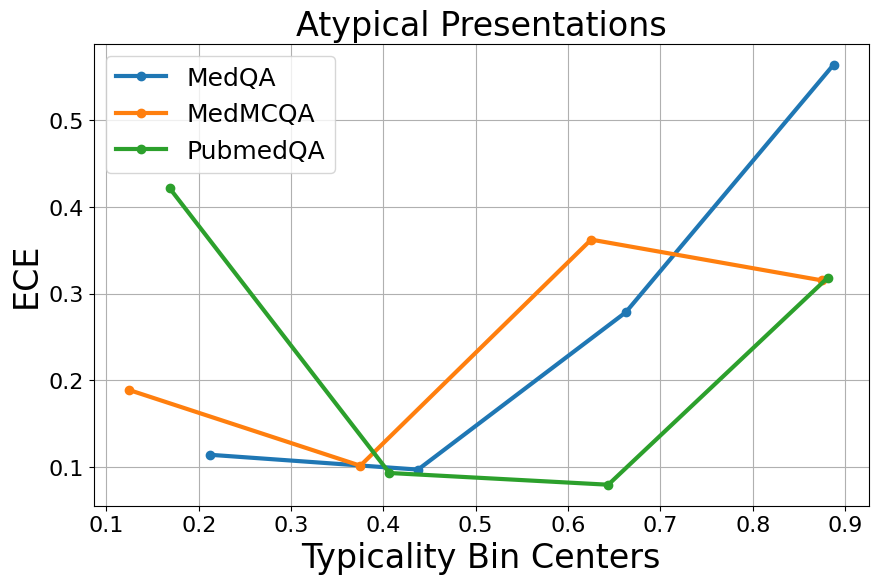} \hfill
  \includegraphics[width=0.48\linewidth]{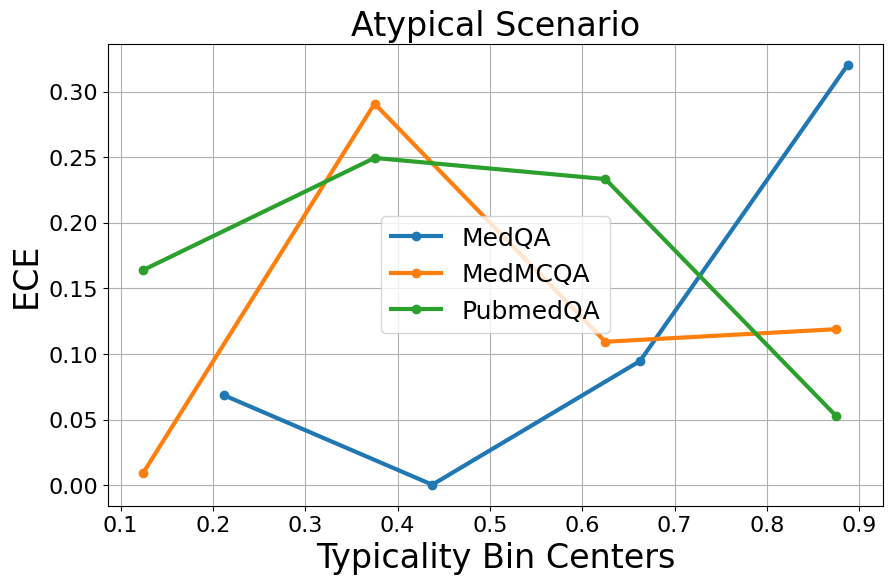}
  \caption {ECE by Typicality bins of GPT4-turbo for Atypical Presentations Aware Recalibration methods}
  \label{fig:ece_by_bins_gpt4}
\end{figure*}

\begin{figure*}[ht]
  \includegraphics[width=0.48\linewidth]{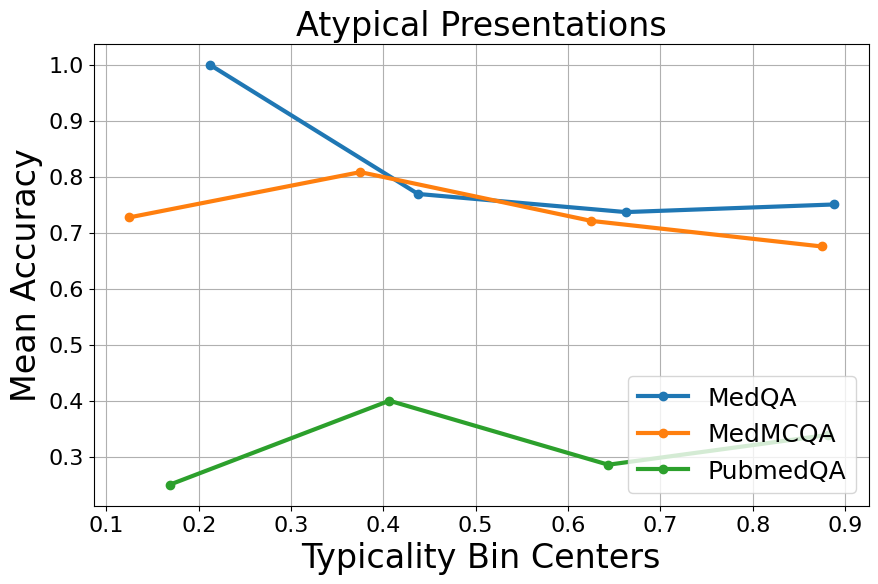} \hfill
  \includegraphics[width=0.48\linewidth]{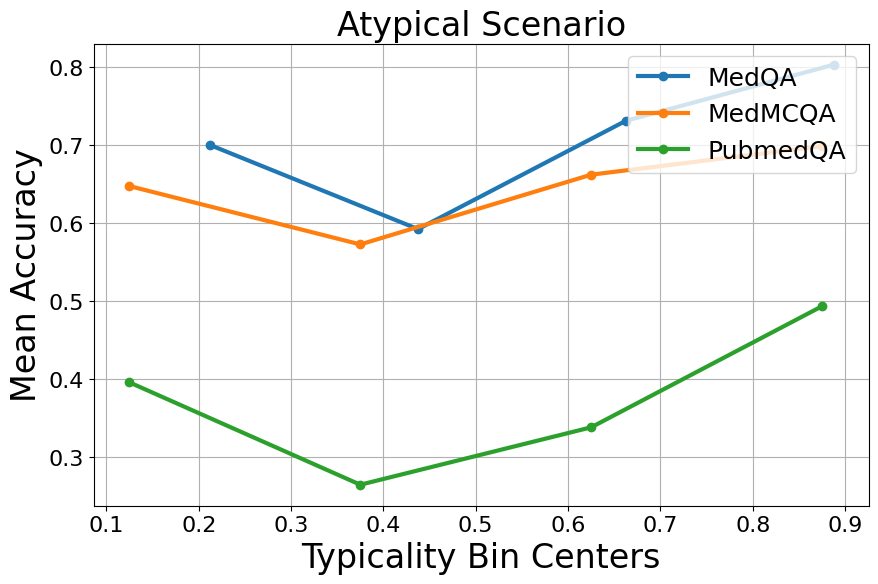}
  \caption {Accuracy by Typicality bins of GPT4-turbo for Atypical Presentations Aware Recalibration methods}
  \label{fig:atypicality_bins_behaviour_gpt4}
\end{figure*}

\begin{figure*}[ht]
  \includegraphics[width=0.32\linewidth]{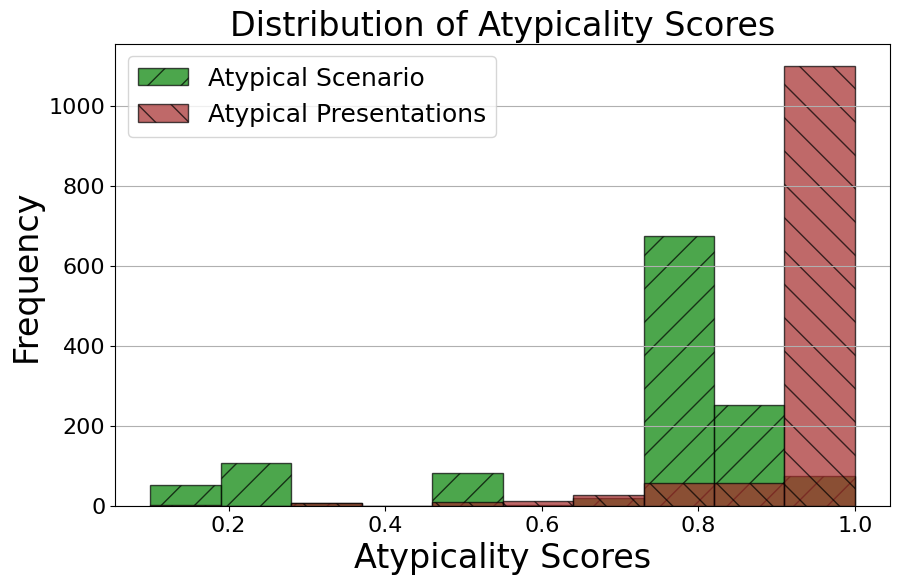} \hfill
  \includegraphics[width=0.32\linewidth]{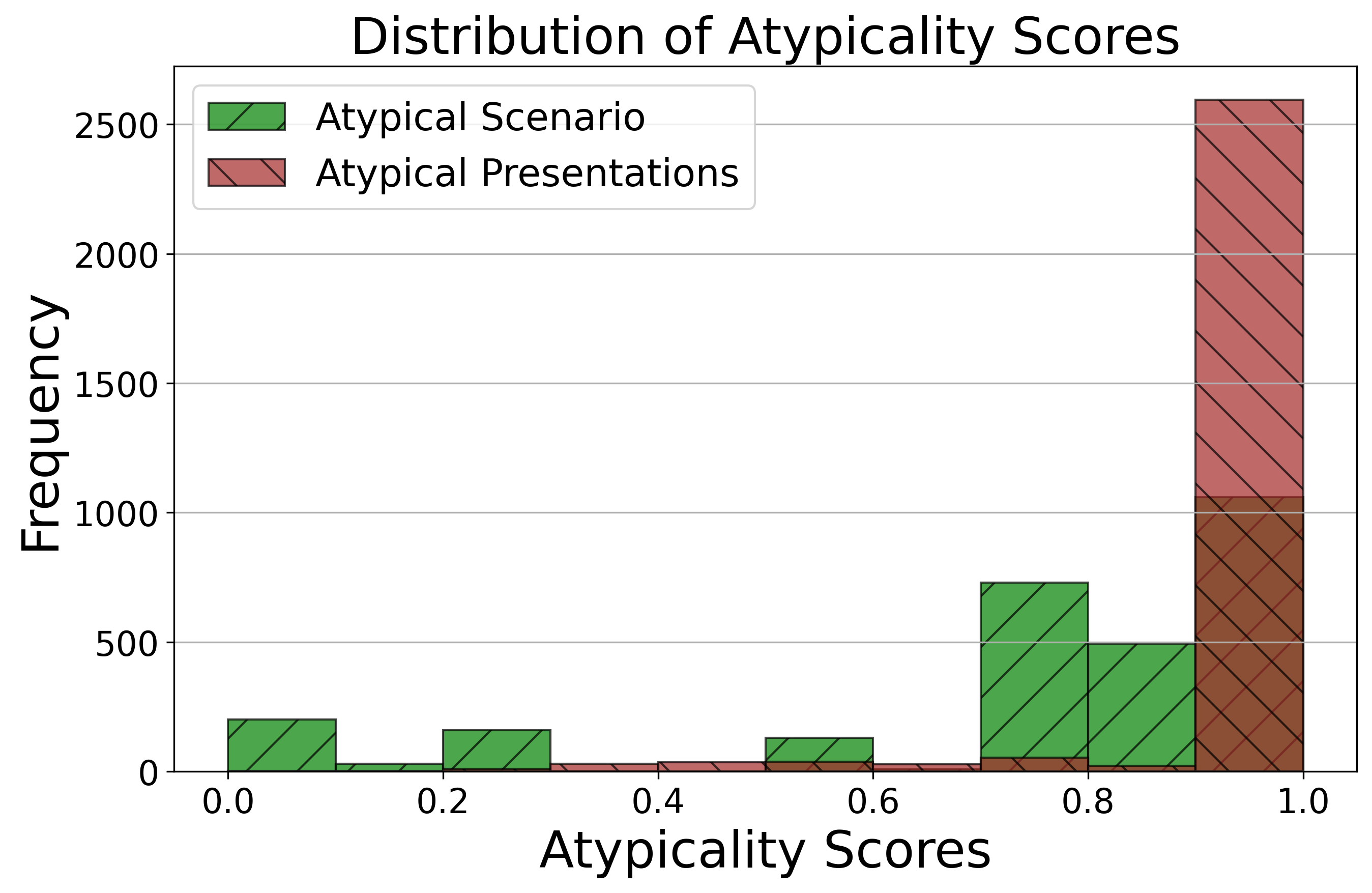}
  \includegraphics[width=0.32\linewidth]{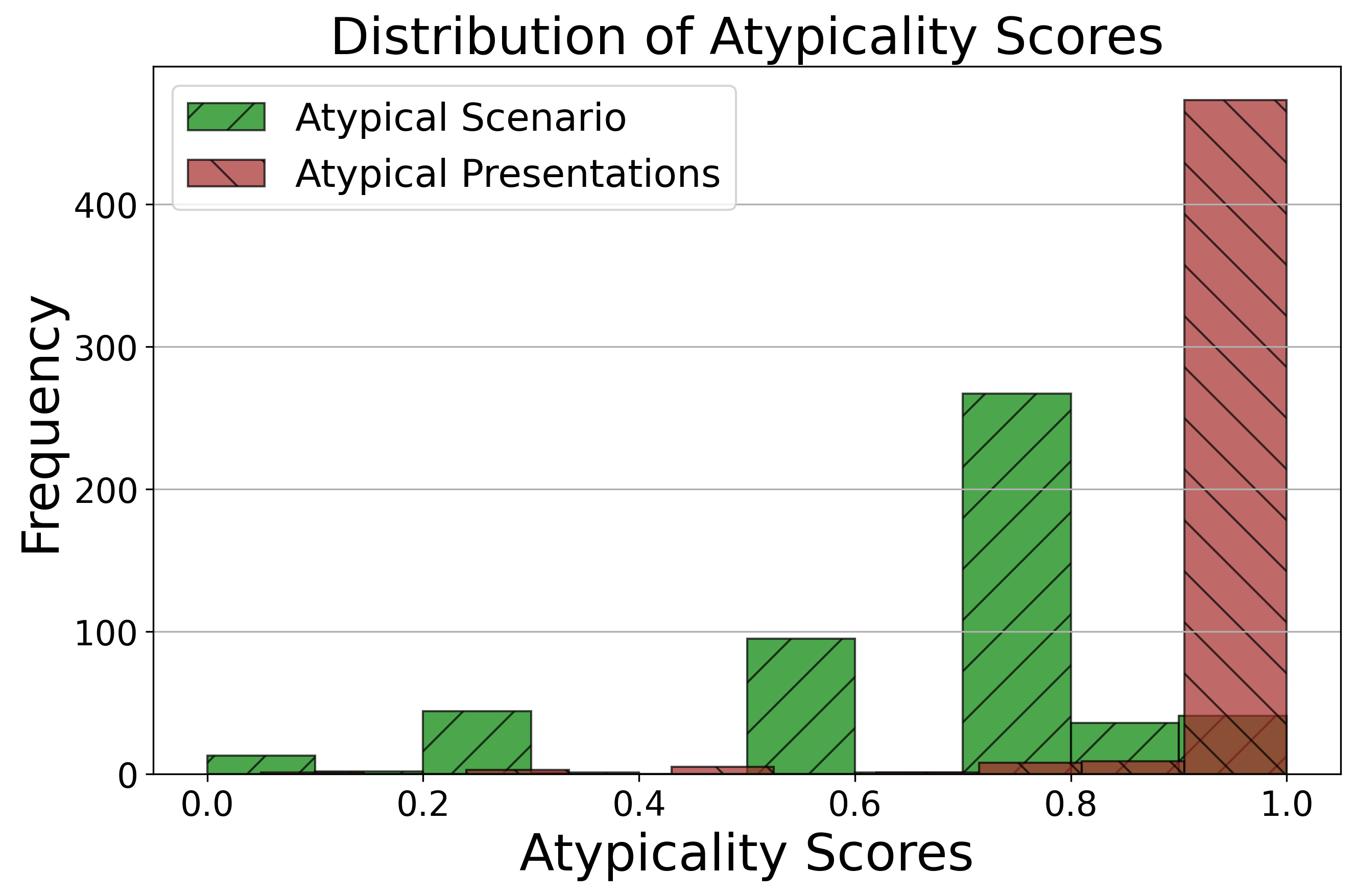}
  \caption {Atypicality Distribution of GPT4}
  \label{fig:atypicality_dist_gpt4}
\end{figure*}

\begin{figure*}[ht]
  \includegraphics[width=0.32\linewidth]{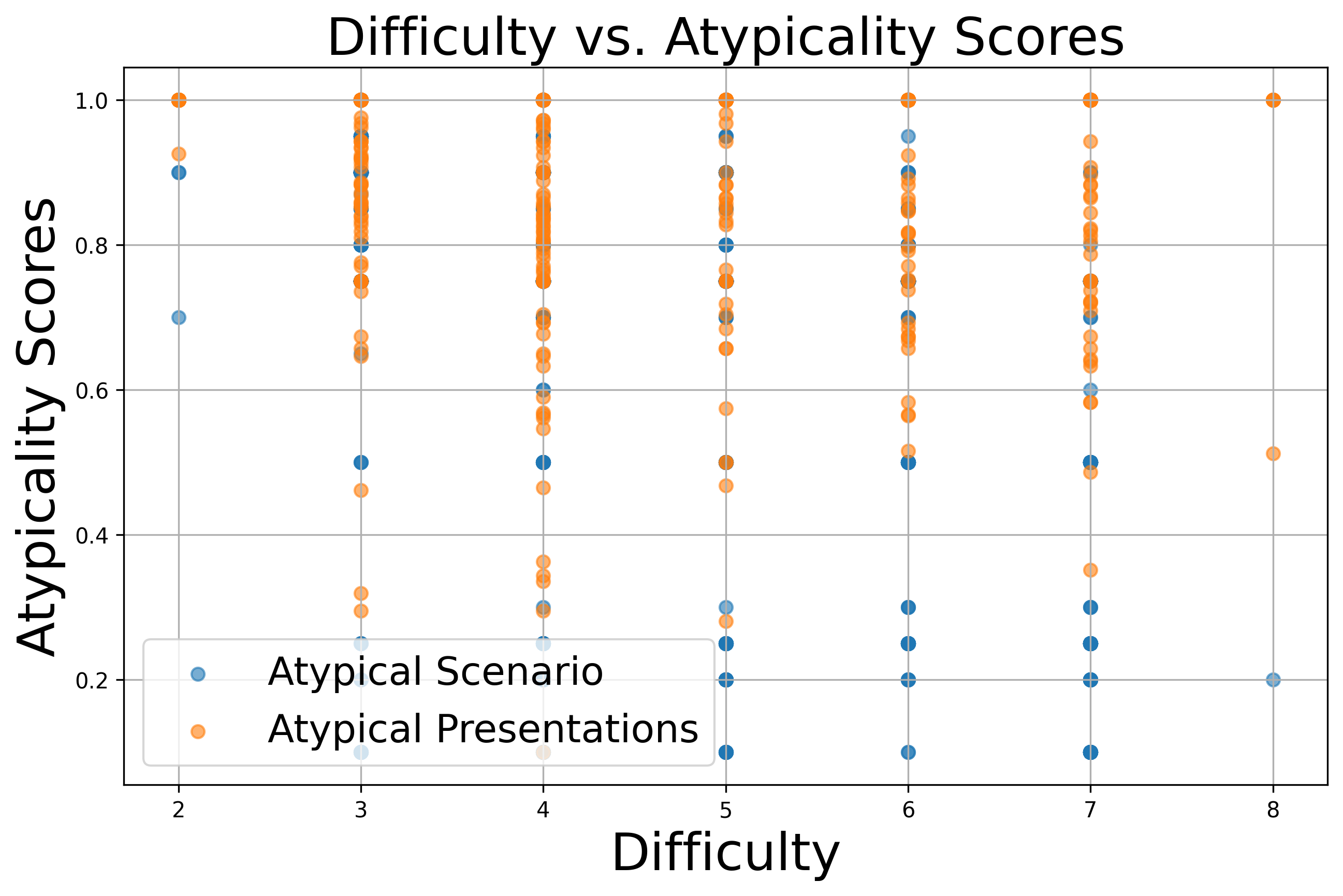} \hfill
  \includegraphics[width=0.32\linewidth]{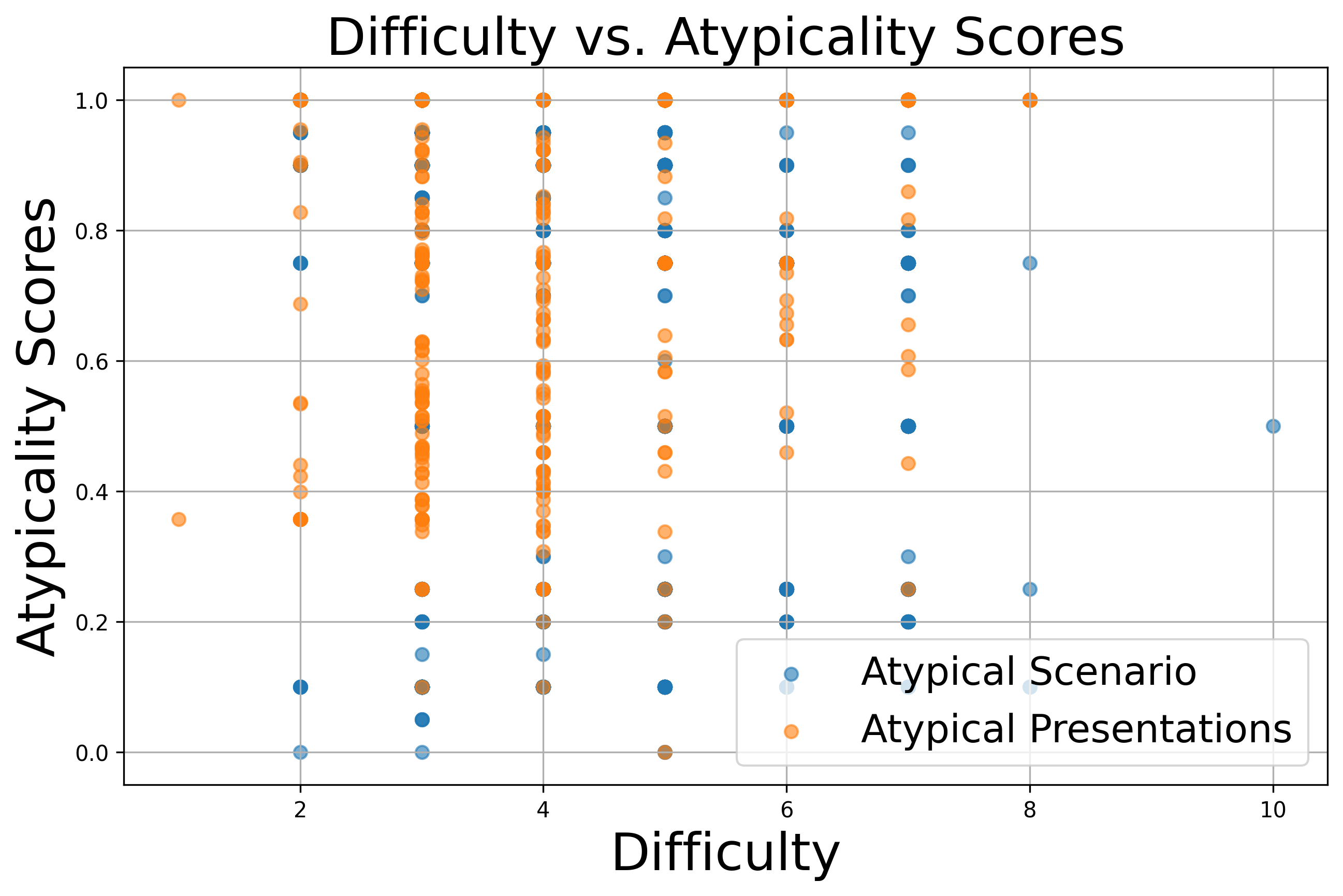}
  \includegraphics[width=0.32\linewidth]{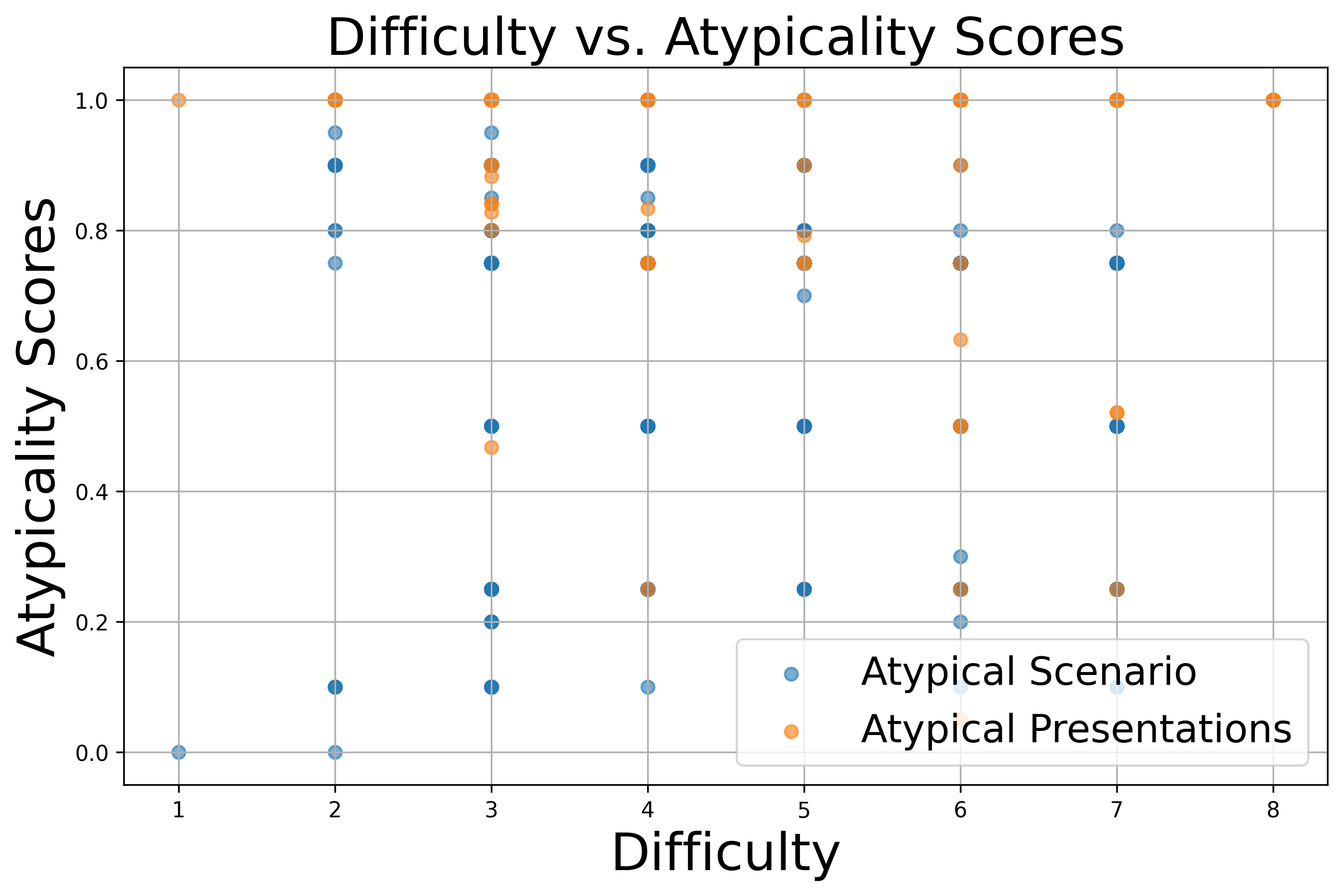}
  \caption {Atypicality by Difficulty of GPT4}
  \label{fig:diff_dist_gpt4}
\end{figure*}

\begin{figure*}[ht]
  \includegraphics[width=1\linewidth]{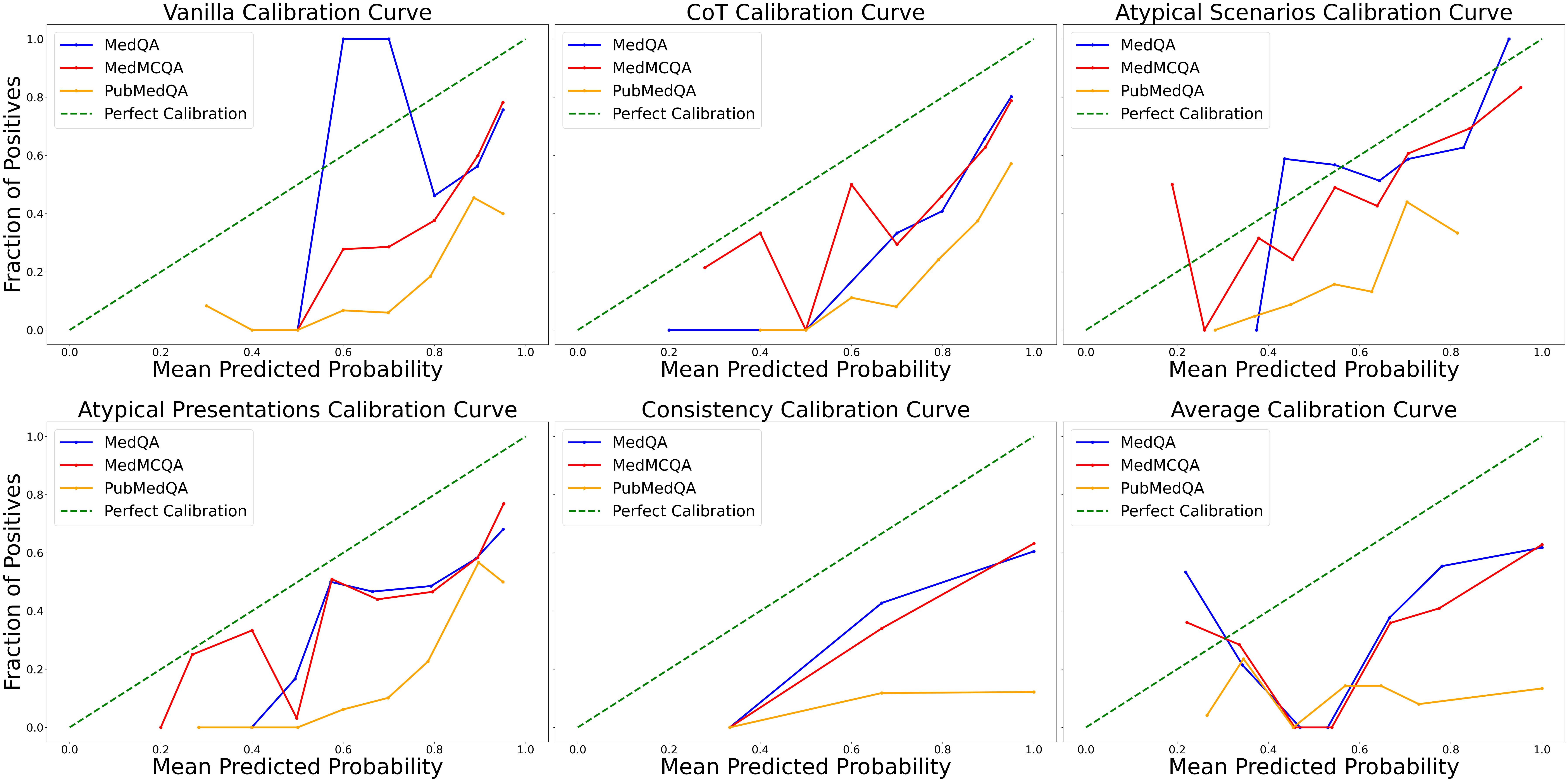} \hfill
  \caption {Claude3-sonnet Calibration curves across all methods}
  \label{fig:calibration_curves_claude}
\end{figure*}

\begin{figure*}[ht]
  \includegraphics[width=0.48\linewidth]{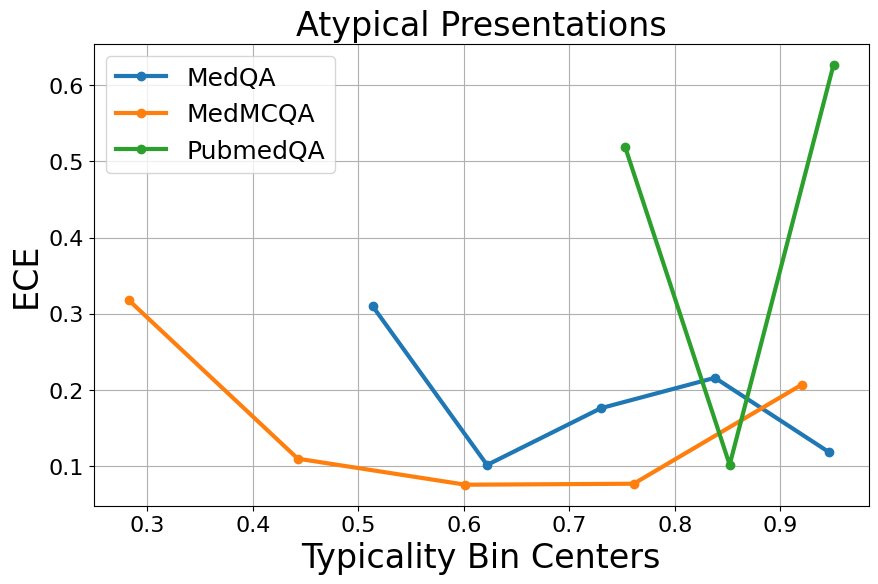} \hfill
  \includegraphics[width=0.48\linewidth]{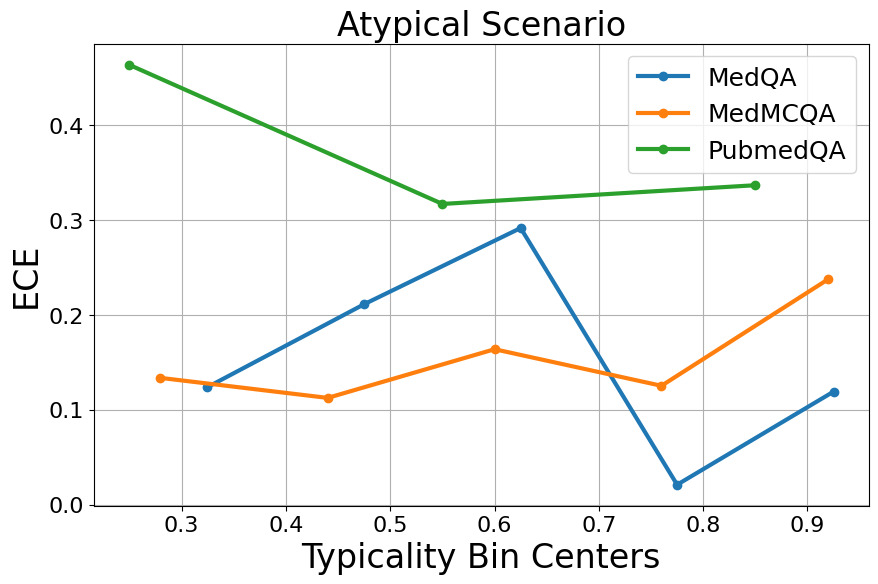}
  \caption {ECE by Typicality bins of Claude3-sonnet for Atypical Presentations Aware Recalibration methods}
  \label{fig:ece_by_bins_claude}
\end{figure*}

\begin{figure*}[ht]
  \includegraphics[width=0.48\linewidth]{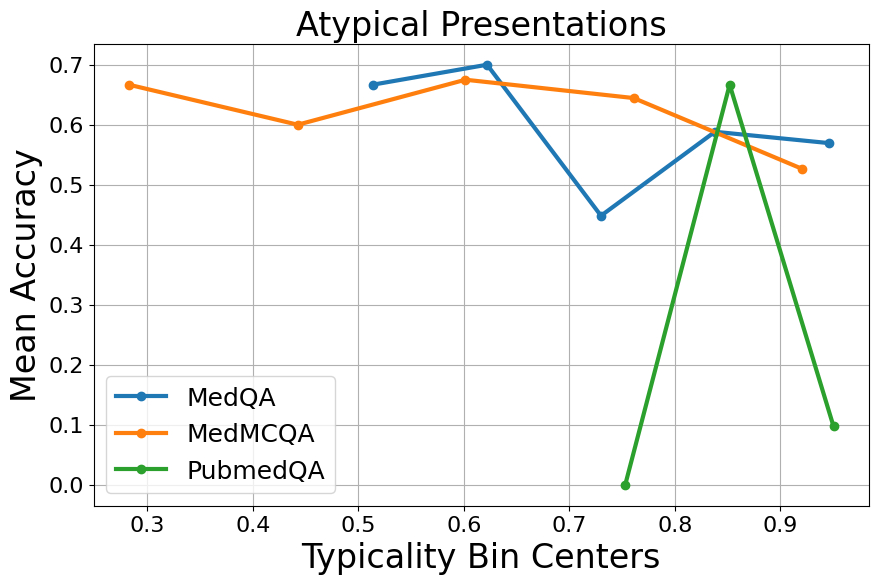} \hfill
  \includegraphics[width=0.48\linewidth]{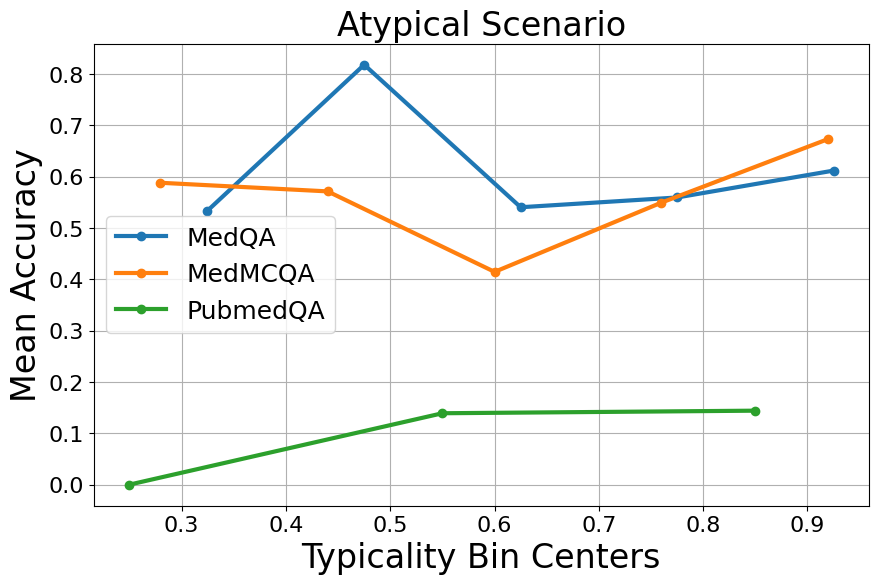}
  \caption {Accuracy by Typicality bins of Claude3-sonnet for Atypical Presentations Aware Recalibration methods}
  \label{fig:atypicality_bins_behaviour_claude}
\end{figure*}

\begin{figure*}[ht]
  \includegraphics[width=0.32\linewidth]{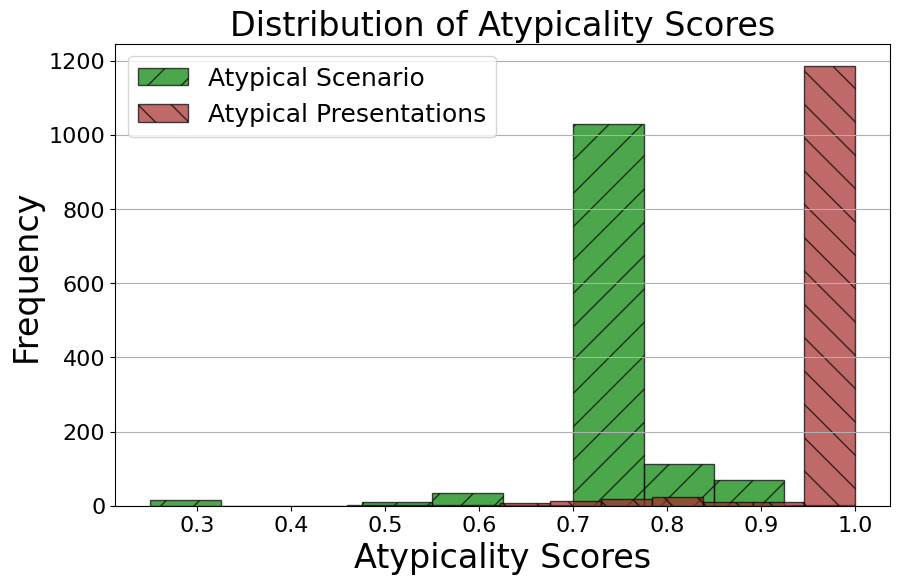} \hfill
  \includegraphics[width=0.32\linewidth]{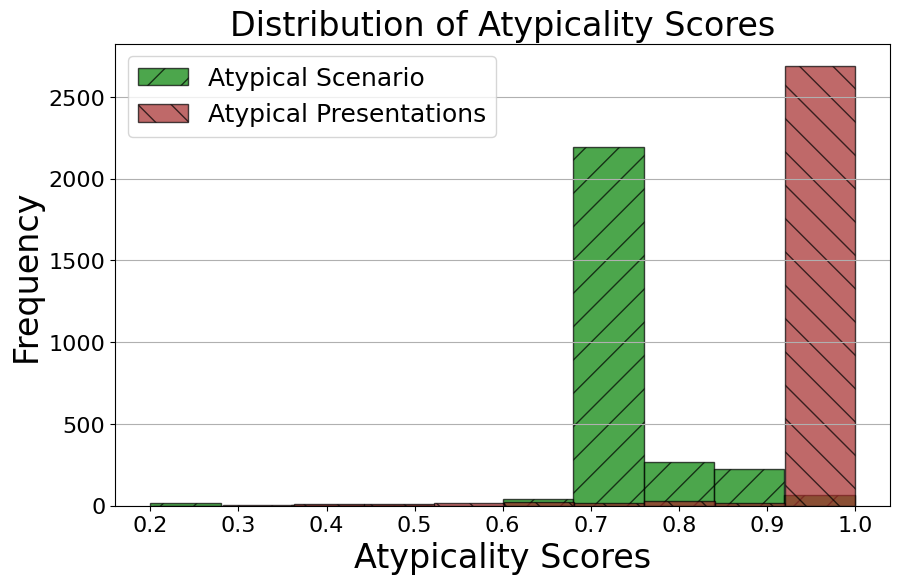}
  \includegraphics[width=0.32\linewidth]{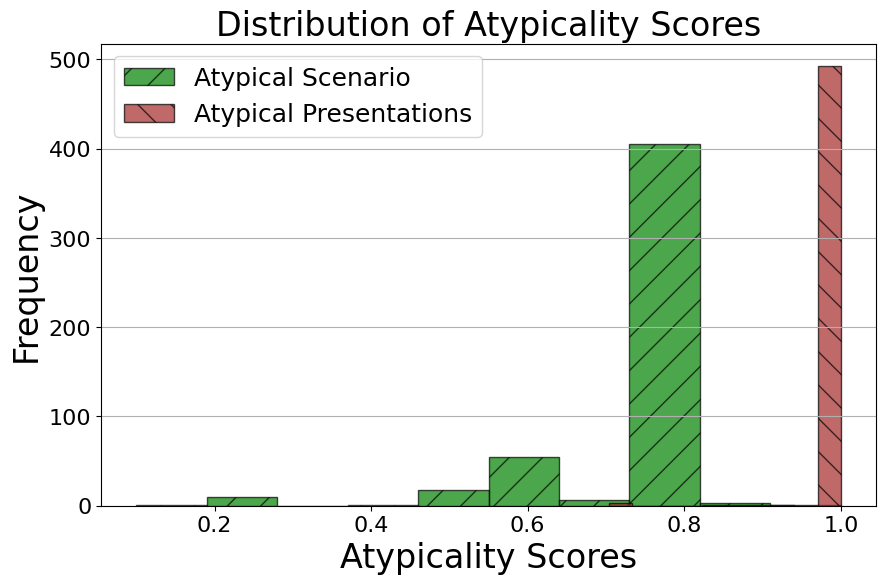}
  \caption {Atypicality Distribution of Claude}
  \label{fig:atypicality_dist_claude}
\end{figure*}

\begin{figure*}[ht]
  \includegraphics[width=0.32\linewidth]{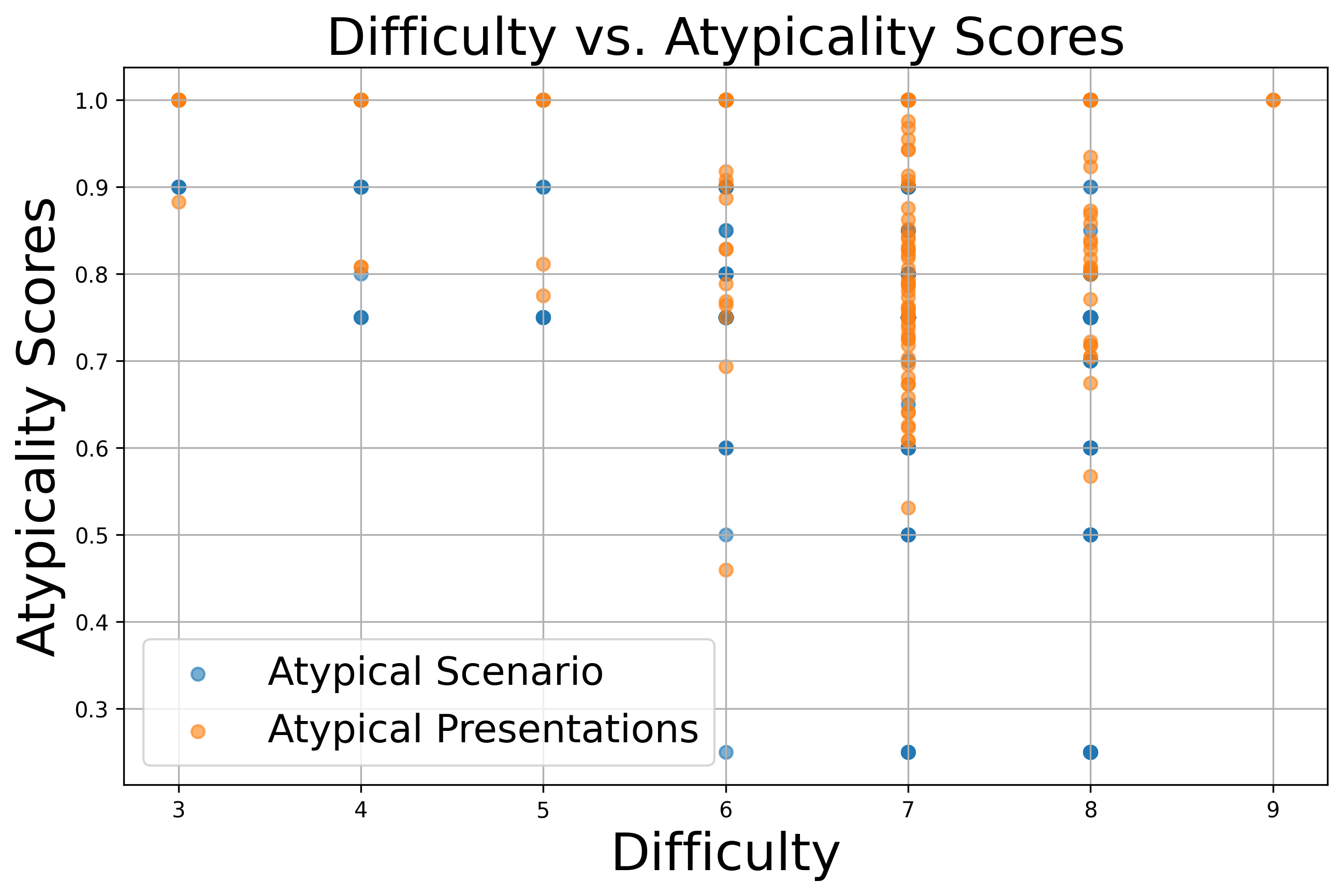} \hfill
  \includegraphics[width=0.32\linewidth]{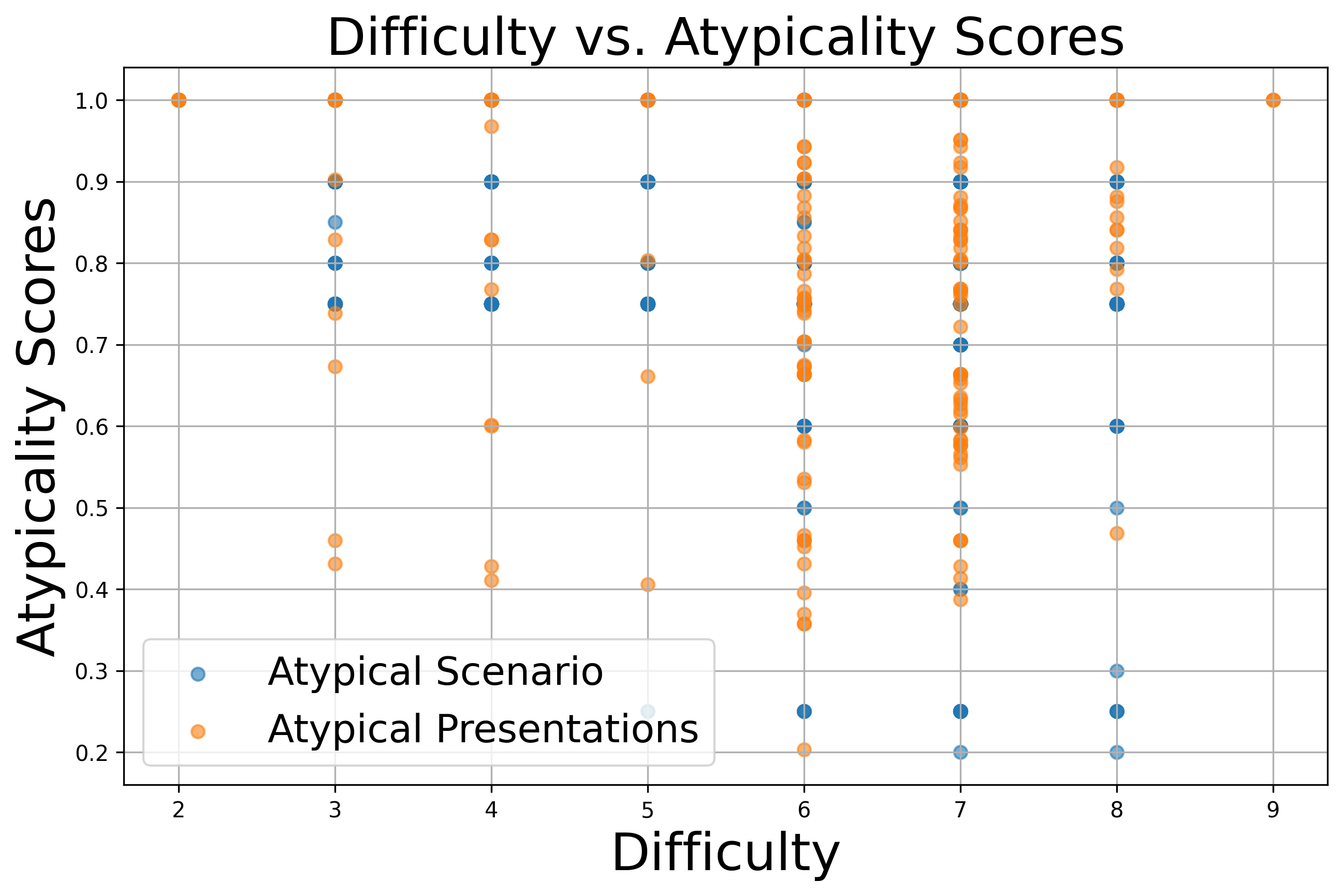}
  \includegraphics[width=0.32\linewidth]{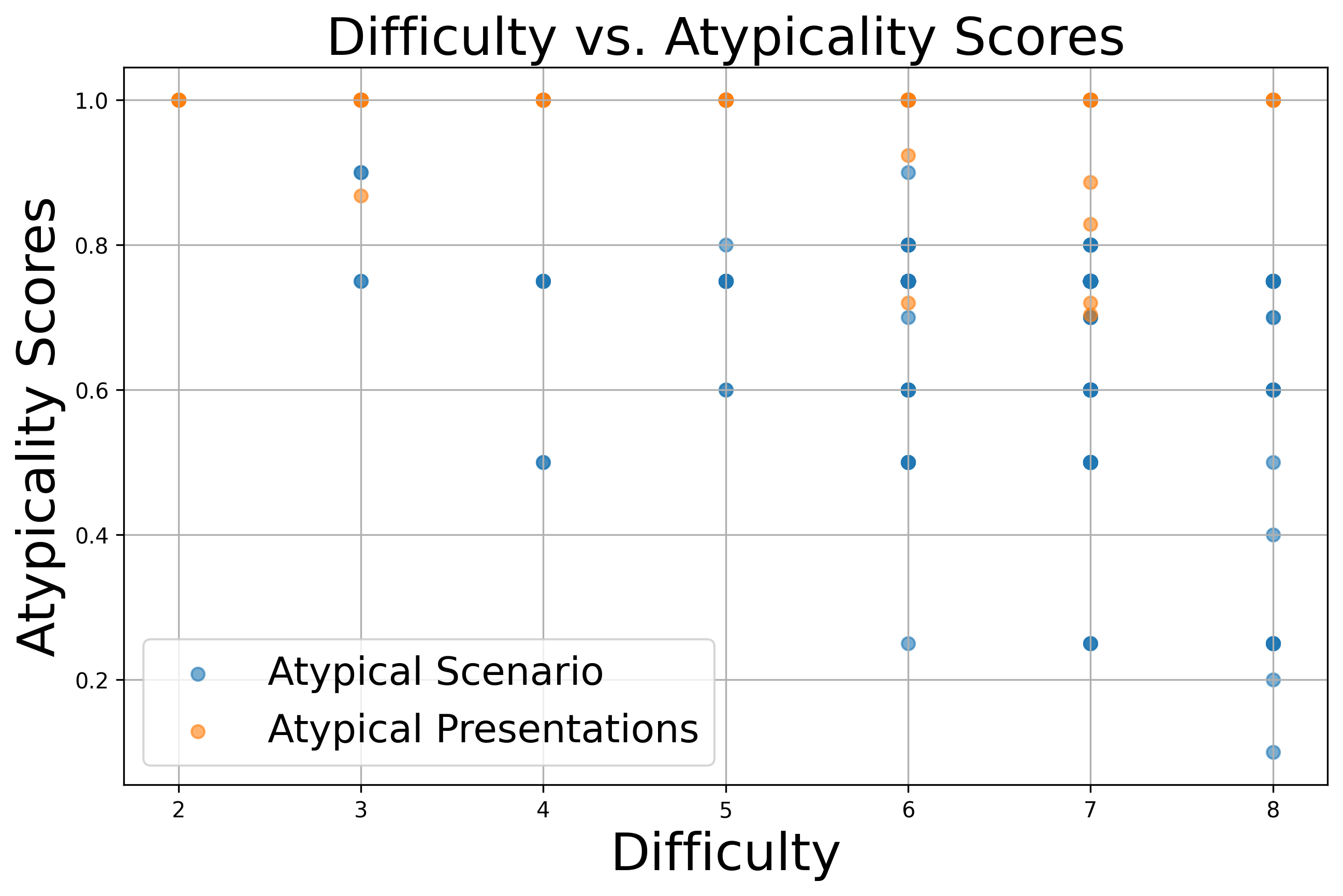}
  \caption {Atypicality by Difficulty of Claude}
  \label{fig:diff_dist_claude}
\end{figure*}

\begin{figure*}[ht]
  \includegraphics[width=1\linewidth]{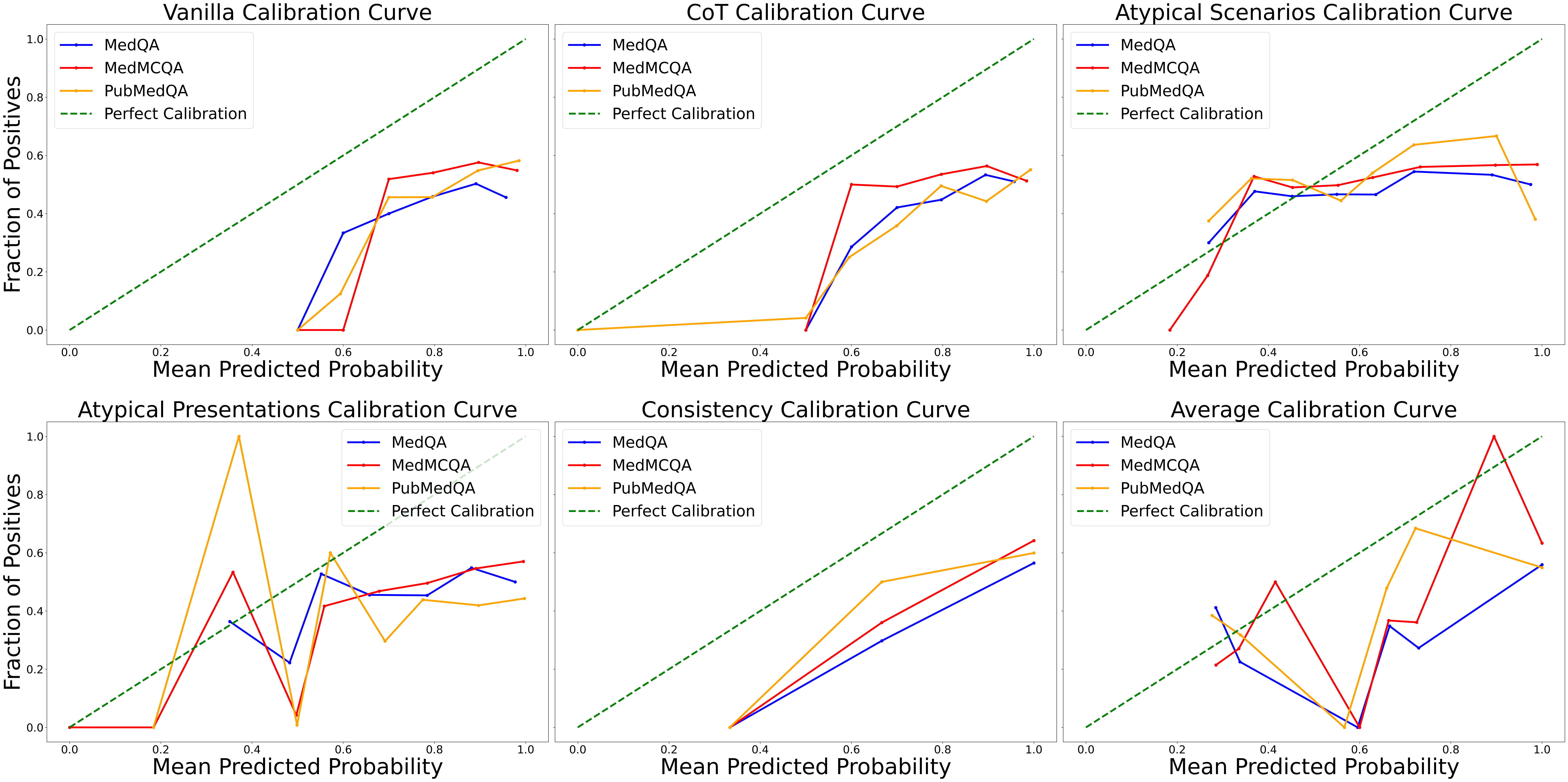} \hfill
  \caption {Gemini Calibration curves across all methods}
  \label{fig:calibration_curves_gemini}
\end{figure*}

\begin{figure*}[ht]
  \includegraphics[width=0.48\linewidth]{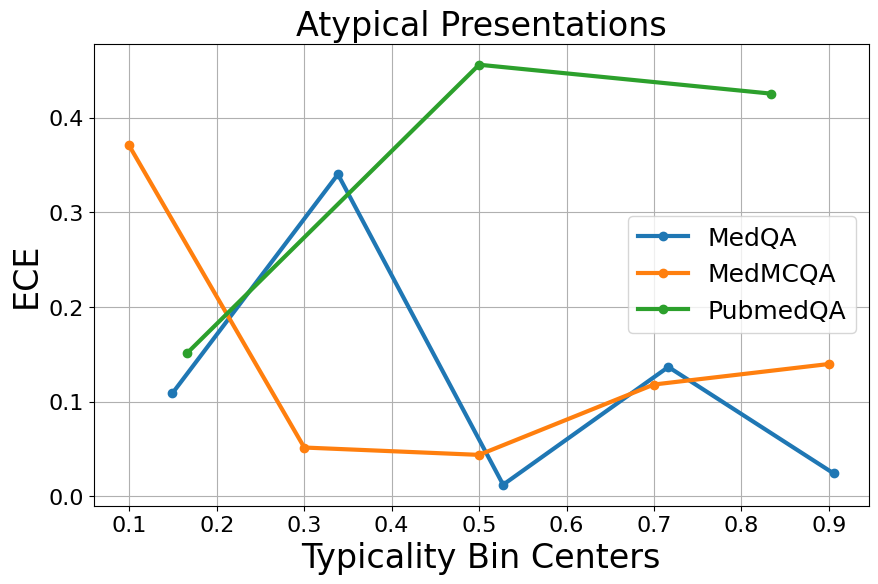} \hfill
  \includegraphics[width=0.48\linewidth]{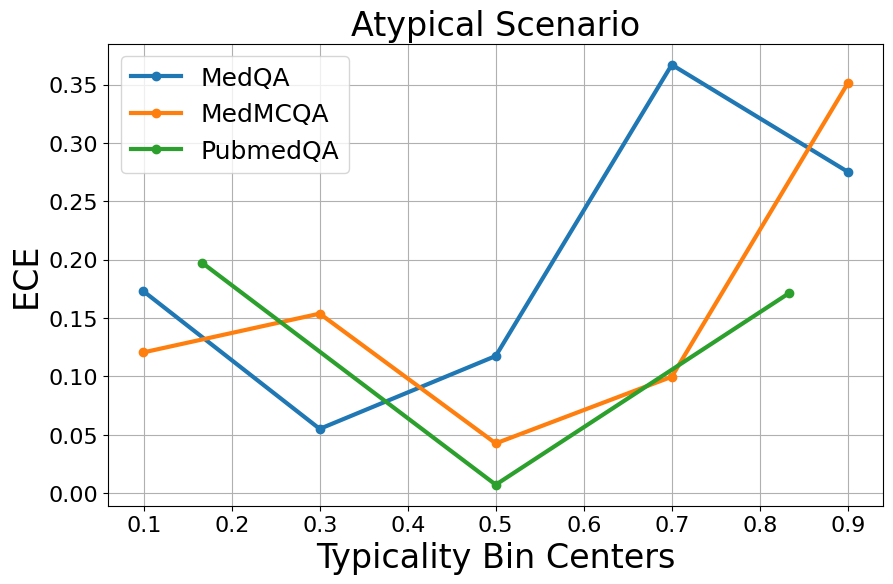}
  \caption {ECE by Typicality bins of Gemini for Atypical Presentations Aware Recalibration methods}
  \label{fig:ece_by_bins_gemini}
\end{figure*}

\begin{figure*}[ht]
  \includegraphics[width=0.48\linewidth]{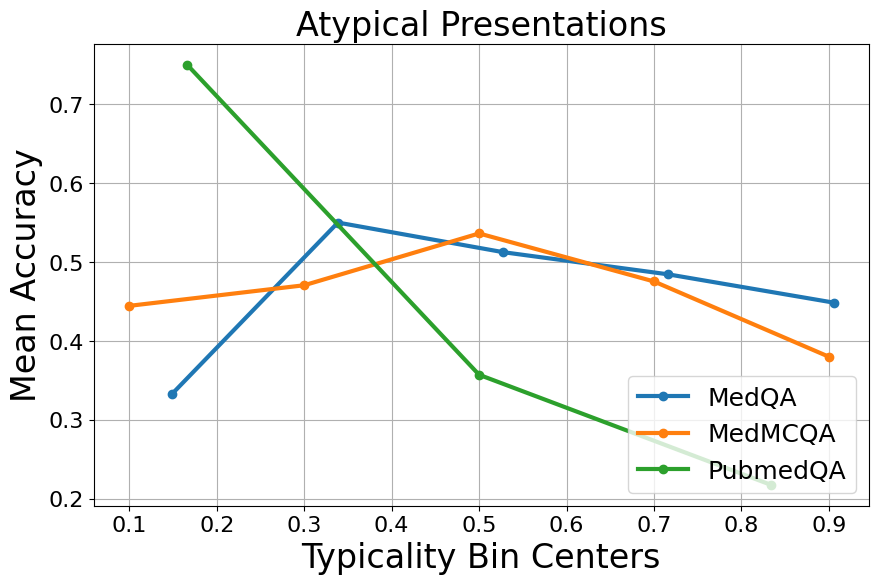} \hfill
  \includegraphics[width=0.48\linewidth]{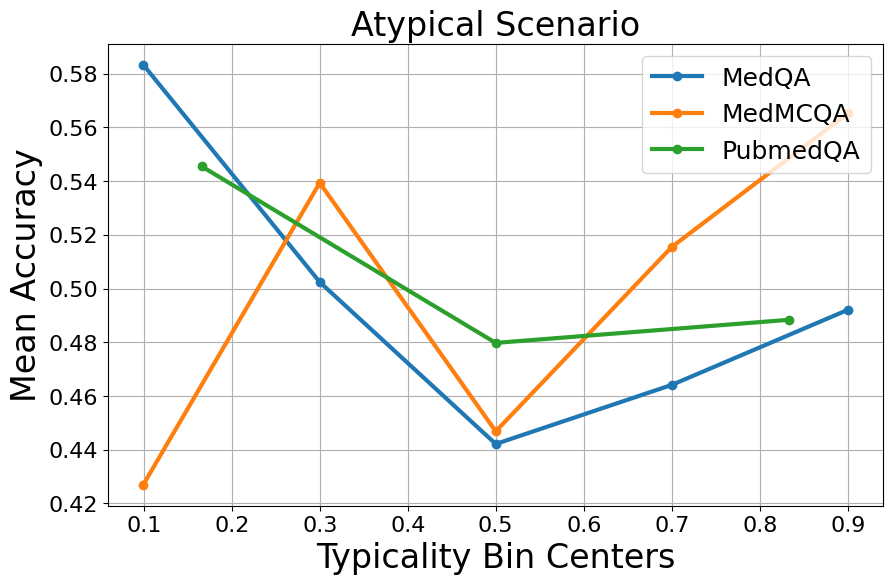}
  \caption {Accuracy by Typicality bins of Gemini for Atypical Presentations Aware Recalibration methods}
  \label{fig:atypicality_bins_behaviour_gemini}
\end{figure*}

\begin{figure*}[ht]
  \includegraphics[width=0.32\linewidth]{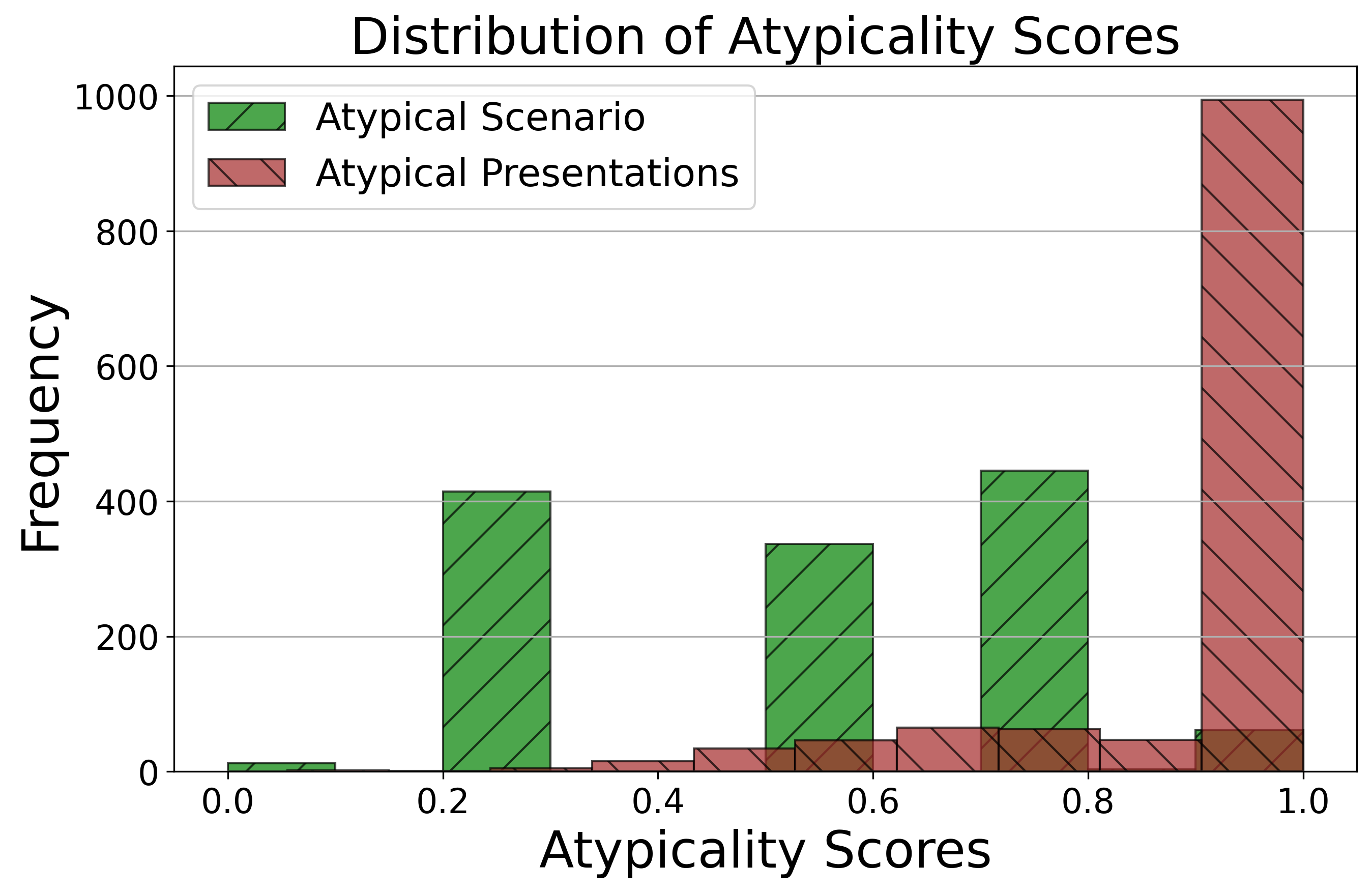} \hfill
  \includegraphics[width=0.32\linewidth]{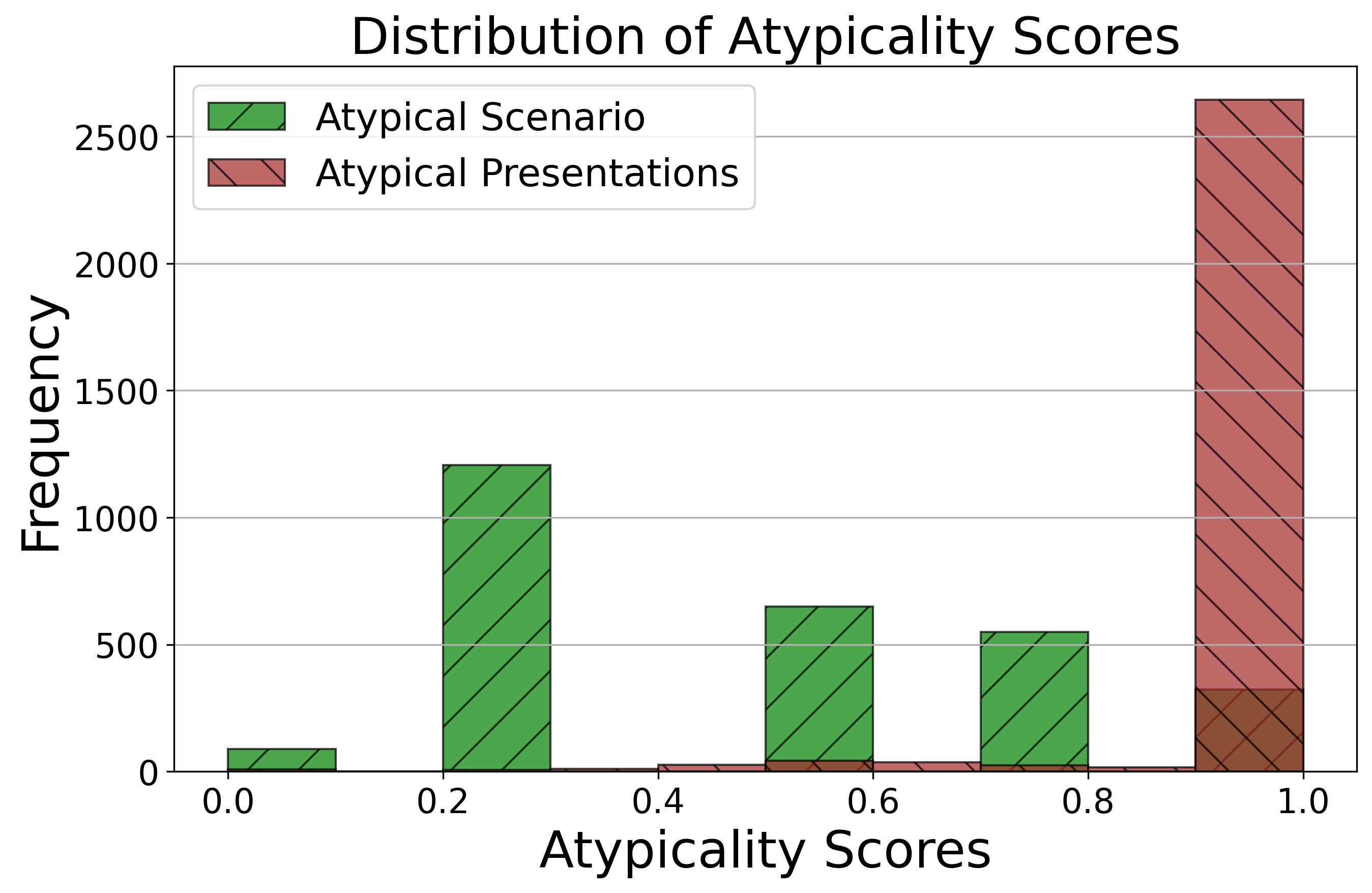}
  \includegraphics[width=0.32\linewidth]{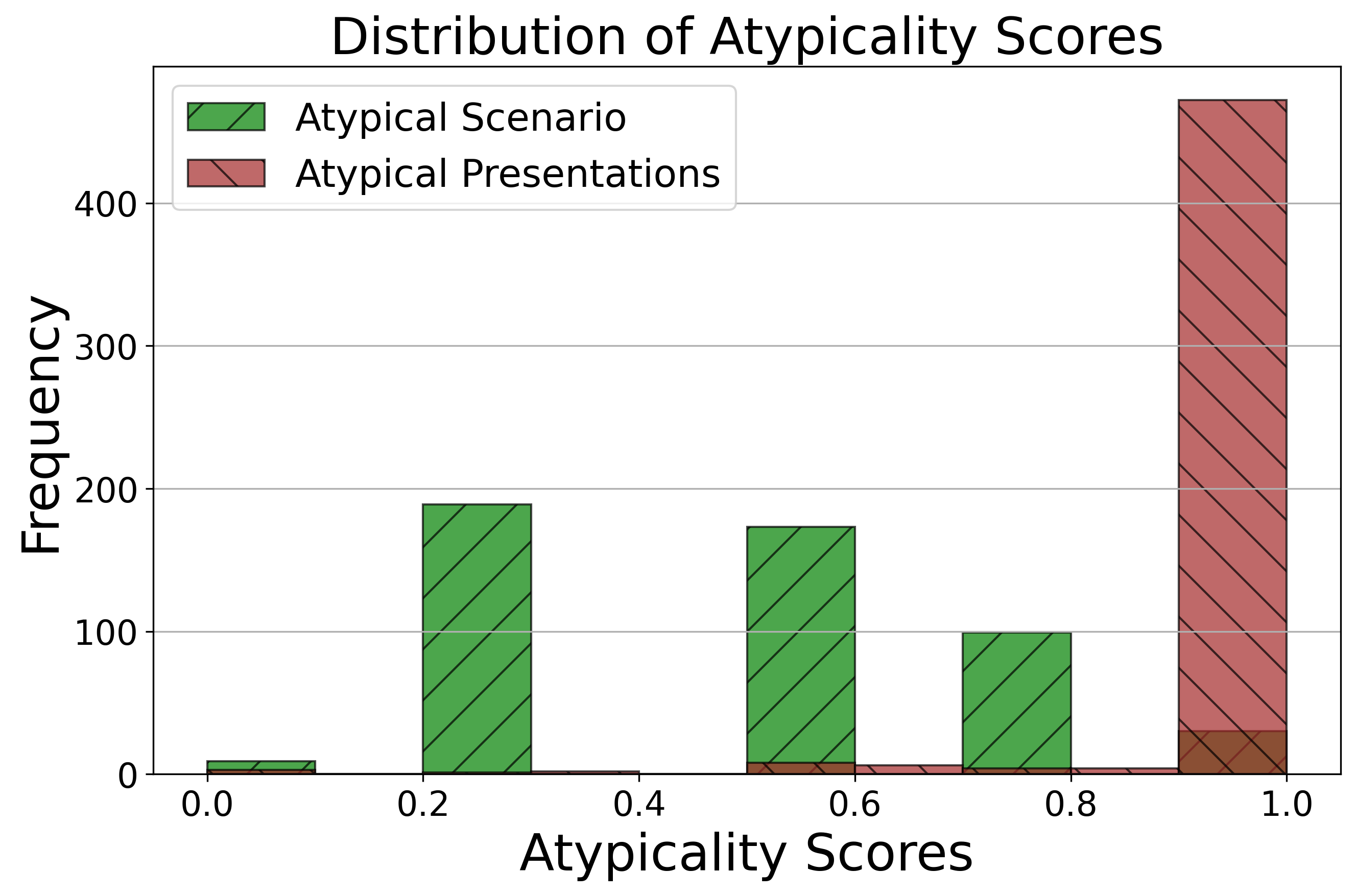}
  \caption {Atypicality Distribution of Gemini}
  \label{fig:atypicality_dist_gemini}
\end{figure*}

\begin{figure*}[ht]
  \includegraphics[width=0.32\linewidth]{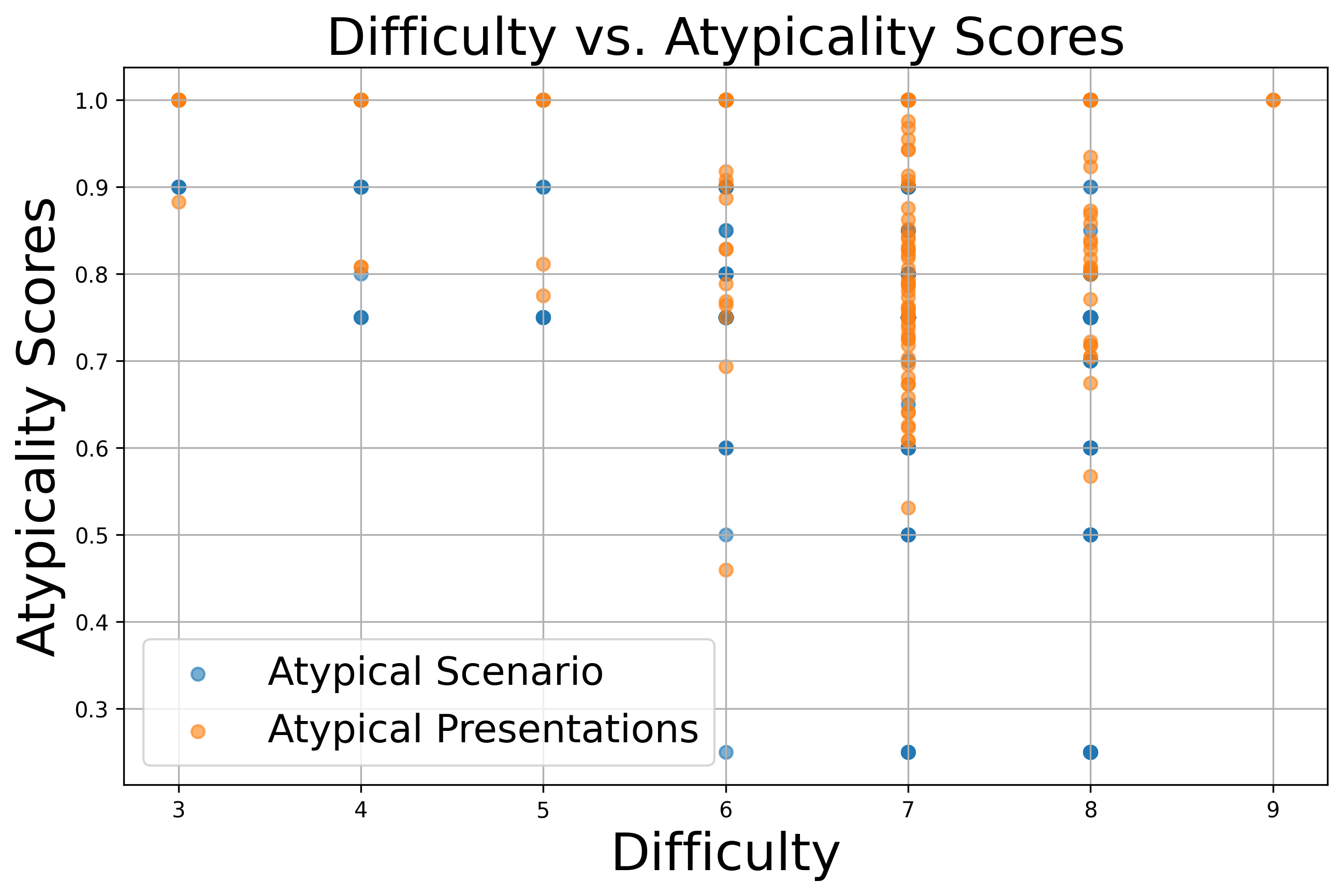} \hfill
  \includegraphics[width=0.32\linewidth]{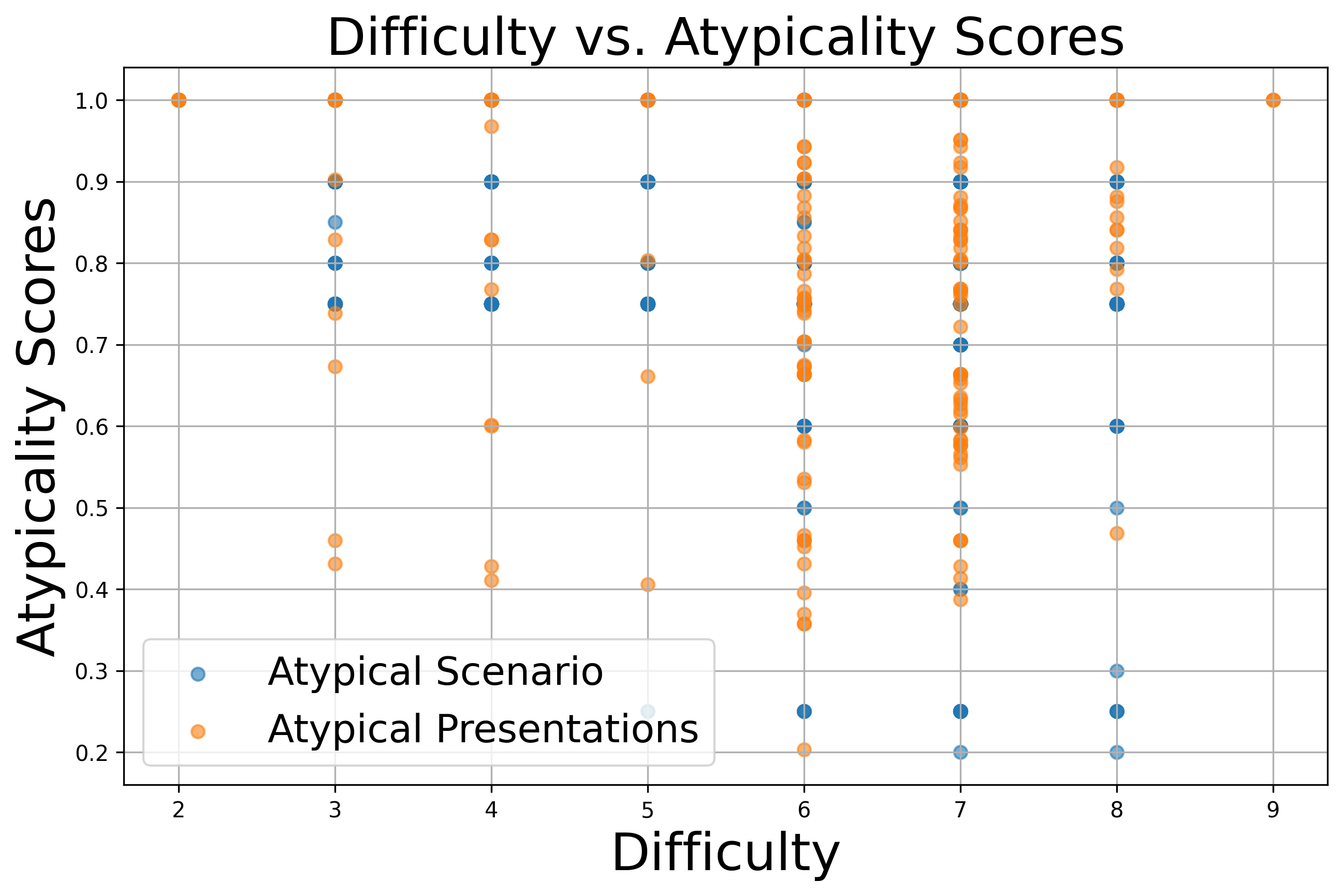}
  \includegraphics[width=0.32\linewidth]{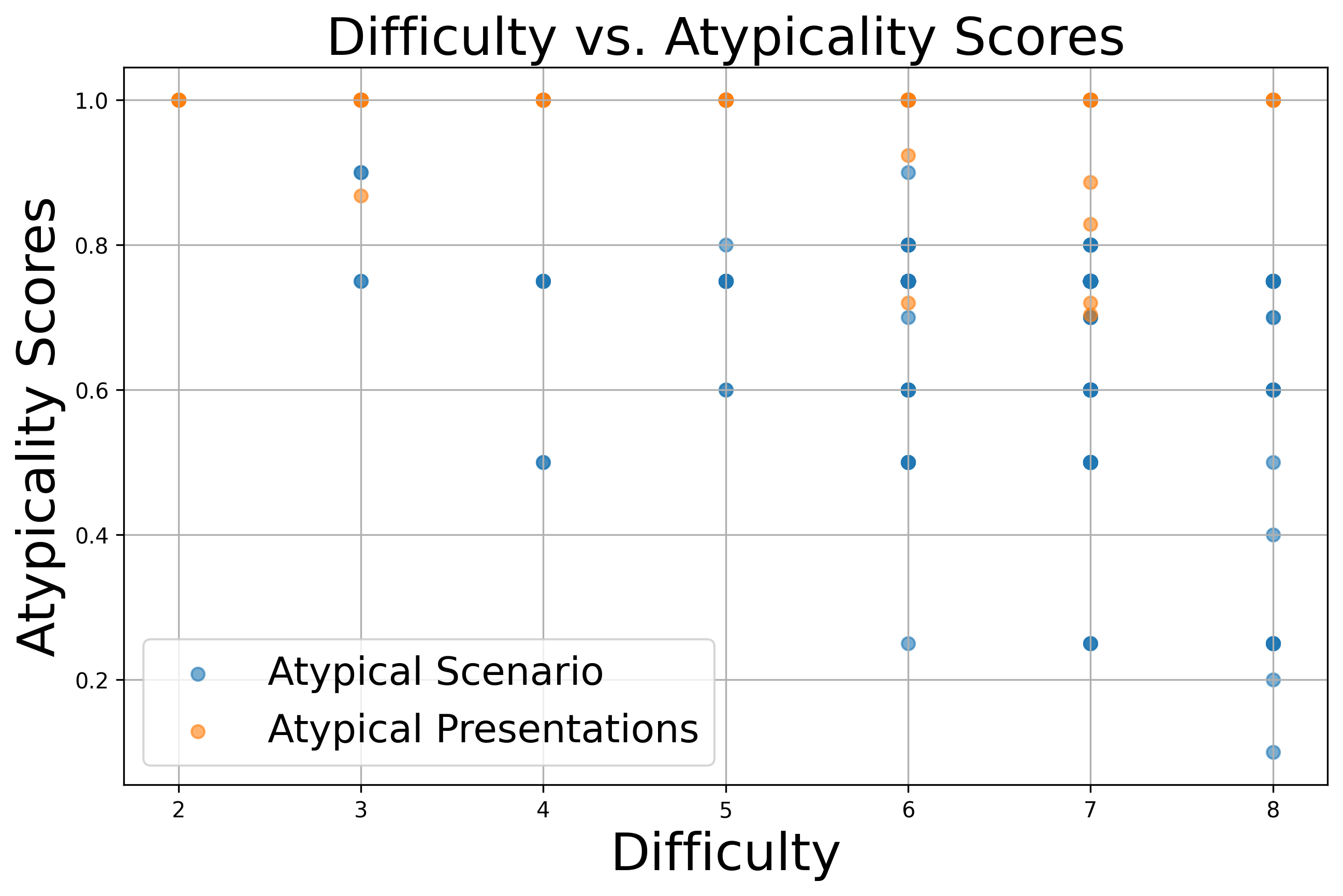}
  \caption {Atypicality by Difficulty of Gemini}
  \label{fig:diff_dist_gemini}
\end{figure*}

\section{Prompt templates}
\label{sec:appendixB}
We provide the full prompt used for \textit{Atypical Scenario} and \textit{Atypical Presentations}. Note that for completeness, the version of prompts provided contains the component of difficulty scores. This component is optional and is only used for  analyzing the relationship between difficulty and atypicality. The prompt templates can be found at Table \ref{tab:full_prompt_templates}.

\begin{table*}[]
\begin{tabular}{|p{0.3\textwidth} p{0.7\textwidth}|}
\hline
\multicolumn{2}{|c|}{\textbf{Prompts}}                                                                                                                                                                                                                                                                                                                                                                                                                                                                                                                                                                                                                                                                                                                                                                                                                                                                                                                                                                    \\ \hline
\multicolumn{1}{|l|}{Atypical Scenario}      & \begin{tabular}[c]{@{}p{0.7\textwidth}@{}}Question and Options: \\ \{question\}\\ \\ First, assess the situation described in the question and assign an atypicality score between 0 and 1, where:\\ - 0 indicates a highly atypical situation, uncommon or rare in such scenarios.\\ - 1 indicates a very typical situation, commonly expected in such scenarios.\\ - Scores between 0 and 1 (such as 0.25, 0.5, 0.75) indicate varying degrees of typicality.\\ \\ Situation Atypicality: {[}Atypicality score{]}\\ \\ Then, provide your response in the following format:\\ Response:\\ - Answer (letter): {[}Letter of the choice{]}\\ - Difficulty: {[}Score on a scale from 1 to 10 with 10 being the hardest{]}\\ - Confidence: {[}Percentage score between 0 and 100\%{]}\\ \\ Answer, Difficulty, and Confidence:\end{tabular}                                                                                                                         \\ \hline
\multicolumn{1}{|l|}{Atypical Presentations} & \begin{tabular}[c]{@{}p{0.7\textwidth}@{}}Question and Options: \\ \{question\}\\ \\ First, assess each symptom and signs with respect to its typicality in the described scenario. Assign an atypicality score between 0 and 1, where:\\ - 0 indicates a highly atypical situation, uncommon or rare in such scenarios.\\ - 1 indicates a very typical situation, commonly expected in such scenarios.\\ - Scores between 0 and 1 (such as 0.25, 0.5, 0.75) indicate varying degrees of typicality.\\ \\ Symptoms and signs:\\ - Symptom 1: {[}Atypical score{]} \\- Symptom 2: {[}Atypical score{]}\\- Symptom 3: {[}Atypical score{]}-\\- ...\\ \\ \\ Then, provide your response in the following format:\\ Response:\\ - Answer (letter): {[}Letter of the choice{]}\\ - Difficulty: {[}Score on a scale from 1 to 10 with 10 being the hardest{]}\\ - Confidence: {[}Percentage score between 0 and 100\%{]}\\ \\ Answer, Difficulty, and Confidence:\end{tabular} \\ \hline
\end{tabular}
\caption{\label{tab:full_prompt_templates}
    Complete prompts used for Atypical Presentations Aware Recalibration framework
  }
\end{table*}

\end{document}